\documentclass[10pt,twocolumn,letterpaper]{article}

\usepackage{cvpr}              
\usepackage{times}
\usepackage{amsmath}
\usepackage{units}
\usepackage{amsmath, amsthm, amssymb}
\usepackage{mathtools}
\usepackage{amssymb}
\usepackage{nicefrac}
\usepackage{xfrac}
\usepackage{bm}
\usepackage{mathdefs}
\usepackage{graphicx}
\usepackage{colortbl}
\usepackage{xspace}
\usepackage{booktabs}
\usepackage{caption}
\usepackage{color}
\usepackage{xcolor}
\usepackage[accsupp]{axessibility} 

\usepackage{comment}
\usepackage{algpseudocode}

\usepackage[linesnumbered,ruled,vlined]{algorithm2e} 

\SetAlFnt{\small}
\SetAlCapFnt{\small}
\SetAlCapNameFnt{\small}
\SetAlCapHSkip{0pt}

\newcommand{\algobreak}{\textbf{break}}

\usepackage[pagebackref=true,breaklinks=true,letterpaper=true,colorlinks,bookmarks=false]{hyperref}

\usepackage[capitalize]{cleveref}
\usepackage{adaptive}

\graphicspath{{figures/}}

\def\cvprPaperID{7228} 
\def\confName{CVPR}
\def\confYear{2022}

\begin{document}

\title{\vspace{-1.0in}Adaptive Gating for Single-Photon 3D Imaging}

\author{\vspace{-.5in} \\
	Ryan Po, Adithya Pediredla, Ioannis Gkioulekas\\
	Robotics Institute, Carnegie Mellon University
}
\maketitle
\begin{abstract}
   \vspace{-0.15in}
   Single-photon avalanche diodes (SPADs) are growing in popularity for depth sensing tasks. However, SPADs still struggle in the presence of high ambient light due to the effects of pile-up. Conventional techniques leverage fixed or asynchronous gating to minimize pile-up effects, but these gating schemes are all non-adaptive, as they are unable to incorporate factors such as scene priors and previous photon detections into their gating strategy. We propose an adaptive gating scheme built upon Thompson sampling. Adaptive gating periodically updates the gate position based on prior photon observations in order to minimize depth errors. Our experiments show that our gating strategy results in significantly reduced depth reconstruction error and acquisition time, even when operating outdoors under strong sunlight conditions.
   \vspace{-0.3in}
\end{abstract}

\section{Introduction}
\label{sec:intro}
\vspace{-0.1in}

Single-photon avalanche diodes (SPADs) are an emerging type of sensor~\cite{Rochas2003SinglePA} that possess single photon sensitivity. Combined with ultrafast pulsed lasers and picosecond-accurate timing electronics, SPADs are becoming increasingly popular in LiDAR systems for 3D sensing applications~\cite{Gupta2019AsynchronousS3, Gupta2019PhotonFloodedS3, Heide2018SubpicosecondP3, Lindell2018Singlephoton3I}. SPAD-based LiDAR is used, e.g., on autonomous vehicles~\cite{2018LiDARDF,RappLidarAV} and consumer devices~\cite{Rangwala2020iPhone}.


Unfortunately, SPAD-based LiDAR faces a fundamental challenge when operating under strong ambient light: background photons due to ambient light can block the detection of signal photons due to the LiDAR laser, an effect know as \emph{pile-up}~ \cite{Coates1968TheCF,Gupta2019AsynchronousS3, Gupta2019PhotonFloodedS3, Becker2015AdvancedTS, Pawlikowska2017SinglephotonTI, Pediredla2018SignalPB}. This effect becomes more pronounced as scene depth increases, and results in potentially very inaccurate depth estimation. 

\begin{figure}[t]
  \includegraphics[width=1\linewidth]{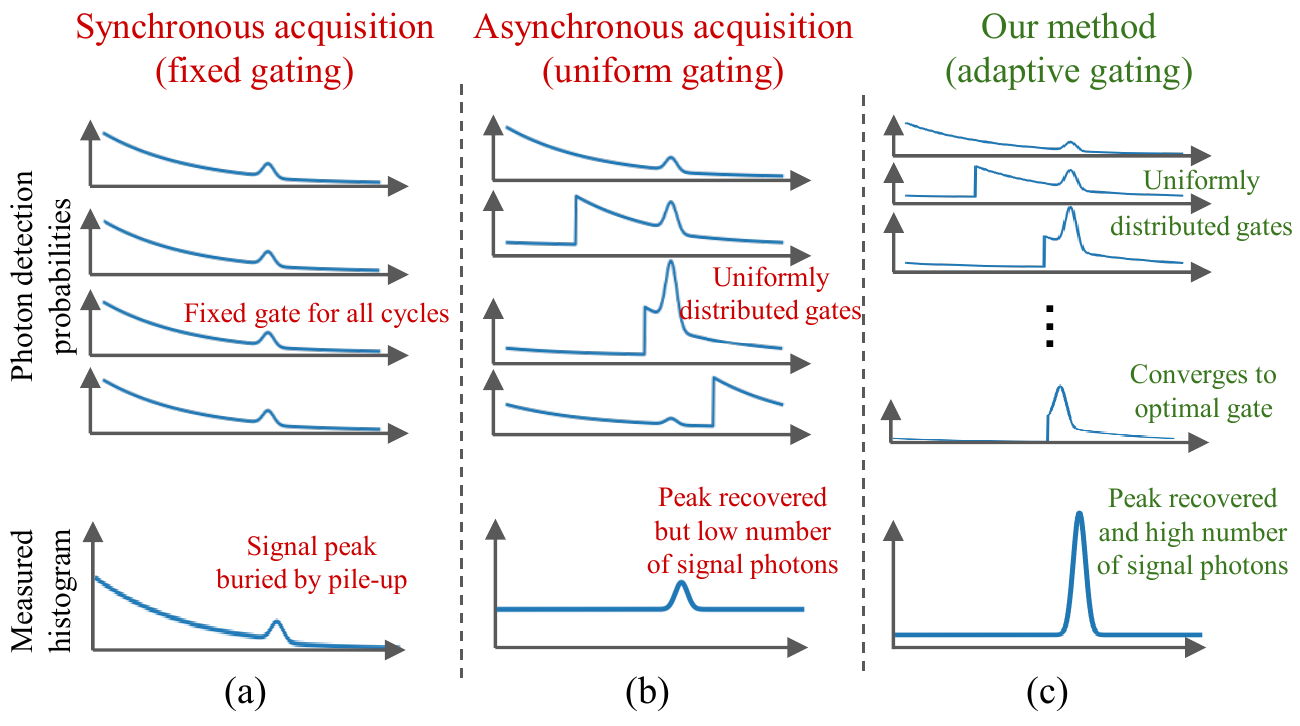}
  \vspace{-0.3in}
  \caption{\textbf{Previous and proposed gating schemes SPAD-based LiDAR.} (a) Under strong ambient light, fixed gating leads to significant pile-up. (b) Uniform gating introduces uniformly-distributed gates and ``averages out'' the effect of pile-up. (c) Adaptive gating introduces a gating scheme that converges to the optimal gate, leading to a large number of detected signal photons.}
  \label{fig:intro}
   \vspace{-0.25in}
\end{figure}

A popular technique for mitigating pile-up is to use \emph{gating mechanisms} that can selectively activate and deactivate the SPAD at specific time intervals relative to laser pulse emissions. Gating can help prevent the detection of early-arriving background photons, and thus favor the detection of late-arriving signal photons. Prior work has proposed different schemes for selecting gating times. A common such scheme is fixed gating, which uses for all laser pulses the same gating time (Figure~\ref{fig:intro}(a)). If this gating time is close to the time-of-flight corresponding to scene depth, then fixed gating greatly increases the detection probability of signal photons, and thus depth estimation accuracy. Unfortunately, it is not always possible to know or approximate the time-of-flight of true depth ahead of time. 

\begin{figure}[t]
	\centering
	\includegraphics[width=1\linewidth]{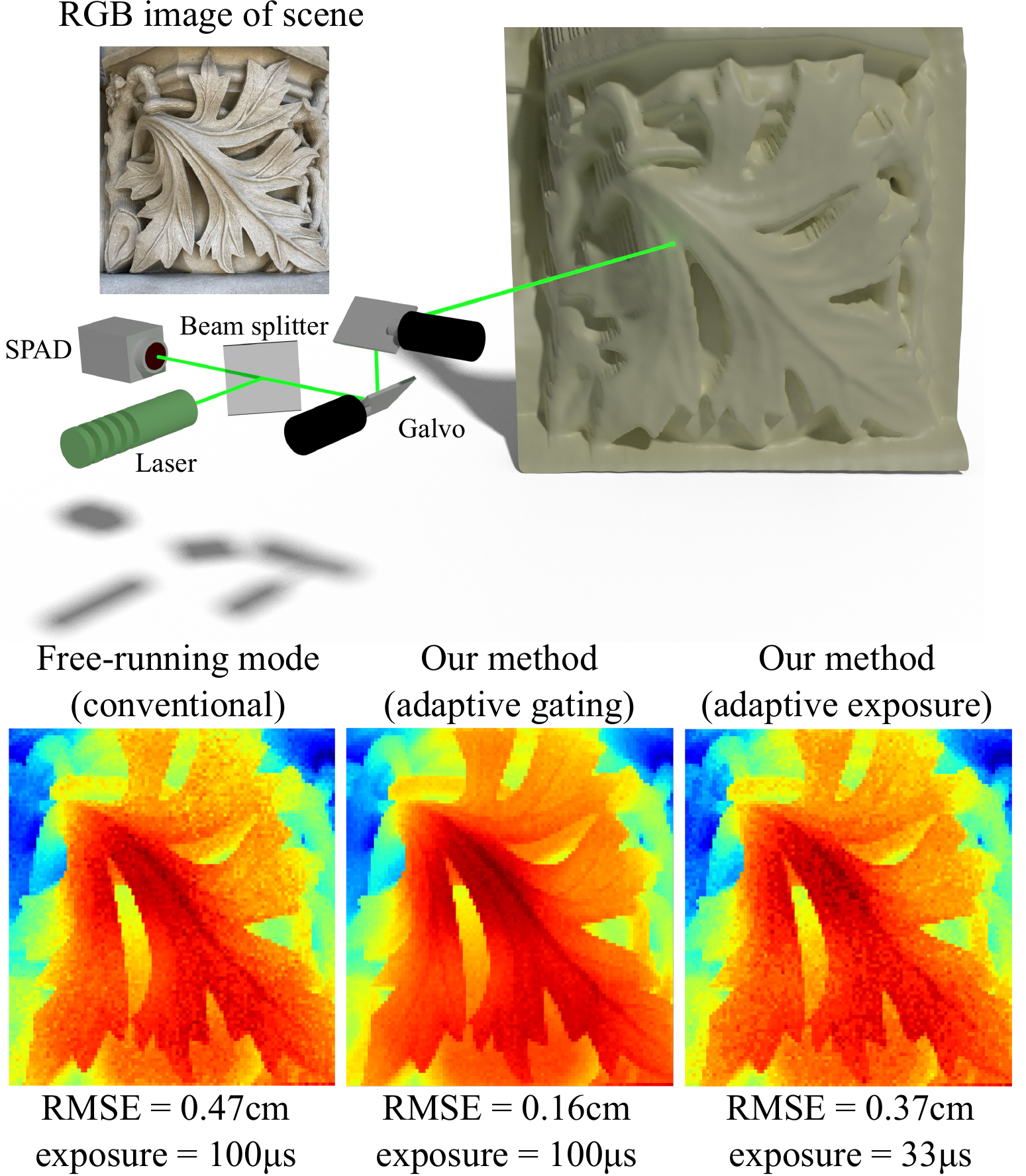}
	\vspace{-0.25in}
	\caption{\textbf{Adaptive gating and adaptive exposure for depth imaging under sunlight.} Adaptive gating reduces depth RMSE $\approx 3\times$ compared to conventional methods for the same acquisition time. When used in conjunction with adaptive exposure, our methods improves frame rate $3\times$ while still achieving lower RMSE.}
	\vspace{-0.3in}  
	\label{fig:teaser}
\end{figure}

More recently, Gupta et al.~\cite{Gupta2019AsynchronousS3} proposed a uniform gating scheme, which uniformly distributes gate times for successive laser pulses across the entire depth range (Figure~\ref{fig:intro}(b)). This helps ``average-out'' the effect of pile-up across all possible depths. Unfortunately, uniform gating does not take into account information about the true scene depth available from either prior knowledge, or from photon detections during previous laser pulses.

We propose a new gating scheme for SPAD-based LiDAR that we term \emph{adaptive gating}. Two main building blocks underlie our gating scheme: First, a probabilistic model for the detection times recorded by SPAD-based LiDAR. Second, the classical Thompson sampling algorithm for sequential experimental design. By combining these two components, our proposed adaptive gating scheme is able to select gating sequences that, at any time during LiDAR acquisition, optimally take advantage of depth information available at that time, either from previous photon detections or from some depth prior. As a useful by-product of our framework, we introduce a variant of our adaptive gating scheme that additionally adapts exposure time, as necessary to achieve some target depth accuracy. We build a SPAD-based LiDAR prototype, and perform experiments both indoors, and outdoors under strong sunlight. Our experiments show that, compared to previous gating schemes, our adaptive gating scheme can reduce either depth estimation error or acquisition time (or both) by more than $50\%$ (Figure~\ref{fig:teaser}), and can take advantage of prior depth information from spatial regularization or RGB images. To ensure reproducibility and facilitate follow-up research, we provide our code and data on the project website~\cite{ProjectWebsite}.


\vspace{-.1in}
\section{Related work}
\vspace{-.1in}
\label{sec:related}

\boldstart{Post-processing for pile-up compensation.} There is extensive prior work on post-processing techniques for compensating the effects of pile-up. Perhaps the best known is Coates' technique~\cite{Coates1968TheCF}, and its generalizations~\cite{Pediredla2018SignalPB, Gupta2019AsynchronousS3,Gupta2019PhotonFloodedS3,Ingle2019HighFP,RappHighFlux,Rapp2019DeadTC}. Coates' technique uses a probabilistic model for photon arrivals to estimate the true incident scene transient, from which we can estimate depth. Alternatively, Heide et al.~\cite{Heide2018SubpicosecondP3} use the same probabilistic model to perform maximum likelihood estimation directly for depth. We discuss how these approaches relate to our work in Section~\ref{sec:background}.

\boldstart{Gating schemes.} Gating refers to the process of desynchronizing laser pulse emission and SPAD acquisition, 
and is commonly used to mitigate the effects of pile-up. The most common gating technique, known as \textit{fixed gating}, uses a fixed delay between laser pulse emission and the start of SPAD acquisition, suppressing the detection of early-arriving photons. If the gating delay approximately matches the scene depth, fixed gating significantly reduces pile-up; otherwise, fixed gating will either have no significant effect, or may even suppress signal photons if the gating delay is after the scene depth. Gupta et al.~\cite{Gupta2019AsynchronousS3} introduced a gating technique that uses uniformly-distributed delays spanning the entire depth range of the SPAD. This \emph{uniform gating} technique helps mitigate pile-up, without requiring approximate knowledge of scene depth. Lastly, Gupta et al.~\cite{Gupta2019AsynchronousS3} showed that it is possible to achieve uniformly-distributed delays between pulse emission and the start of SPAD acquisition by operating the SPAD without gating, at \emph{free-running mode}~\cite{Isbaner2016DeadtimeCO, Cominelli2017HighspeedAL, Acconcia2018FastFF, Rapp2019DeadTC, RappHighFlux}. We discuss fixed gating, uniform gating, and free-running mode in Section~\ref{sec:gating}.


\boldstart{Spatially-adaptive LiDAR.} SPAD-based LiDAR typically uses beam steering to raster-scan individual pixel locations. Recent work has introduced several techniques that, instead of performing a full raster scan, \emph{adaptively} select spatial locations to be scanned, in order to accelerate acquisition~\cite{bergman2020deep,pittaluga2020towards,yamamoto2018efficient}. These techniques are complementary to ours: Whereas they adaptively sample \emph{spatial} scan locations, our technique adaptively samples \emph{temporal} gates.

\boldstart{Other LiDAR technologies.} There are several commercially-available technologies for light detection and ranging (LiDAR), besides SPAD-based LiDAR~\cite{RappLidarAV}. A common alternative in autonomous vehicles uses avalanche photodiodes (APDs)~\cite{li2020lidar}. Compared to SPAD-based LiDAR, APD-based LiDAR does not suffer from pile-up, but has reduced sensitivity. Other LiDAR systems use continuous-wave time-of-flight (CWToF) cameras~\cite{gokturk2004time,whyte2015application,kadambi2016macroscopic}. CWToF-based LiDAR is common in indoor 3D applications~\cite{webb2012beginning,tadano2015depth,anderson2005experimental}. Unlike SPAD-based and APD-based LiDAR, CWToF-based LiDAR is sensitive to global illumination (multi-path interference).

\boldstart{Other SPAD applications.} SPADs find use in biophotonics applications~\cite{bruschini2019single}, including fluorescence-lifetime imaging microscopy~\cite{lakowicz1992fluorescence, bhandari2015blind, barsi2014multi, lee2019coding, van2005fluorescence}, super-resolution microscopy~\cite{madonini2021single}, time-resolved Raman spectroscopy~\cite{madonini2021single}, and time-domain diffuse optical tomography~\cite{zhao2021high, pifferi2016new}. Other applications include non-line-of-sight imaging~\cite{Xin2019ATO, Buttafava2015NonlineofsightIU, OToole2018ConfocalNI, liu2019non}, and high-dynamic-range imaging~\cite{Ingle2019HighFP, ingle2021passive}.

\vspace{-0.11in}
\section{Background on SPAD-based LiDAR}\label{sec:background}
\vspace{-0.1in}

We discuss necessary background on 3D imaging with single-photon avalanche diodes (SPADs). 
We consider single-pixel SPAD-based LiDAR systems with controllable gating. Such a system comprises: a) an ultrafast pulsed laser that can emit short-duration light pulses at a pulse-to-pulse frequency (or \emph{repetition rate}) $\ptpfrequency$; b) a single-pixel SPAD that can detect individual incident photons; c) gating electronics that can activate the SPAD at some controllable time after pulse emissions; and d) time-correlation electronics that time photon detections relative to pulse emissions. We assume that both the gating and time-correlation electronics have the same temporal resolution $\resolution$. Typical orders of magnitude are a few $\unit[]{ps}$ for pulse duration, hundreds of $\unit[]{ps}$ for $\resolution$, and tens of $\unit[]{MHz}$ for $\ptpfrequency$. The laser and SPAD are commonly coaxial, and rely on beam steering (e.g., through a galvo or MEMS mirror) to produce 2D depth estimates. Figure~\ref{fig:teaser} shows a schematic of such a setup.

At each scan point, the LiDAR uses \emph{time-correlated single photon counting} (TCSPC) to estimate depth. This process comprises $\numcycles$ \emph{cycles}, where at each cycle a sequence of steps takes place: At the start of the $\cycle$-th cycle, the laser emits a pulse. Gating activates the SPAD at \emph{gate time} $\gate_\cycle$ after the start of the cycle. The SPAD remains active until it detects the first incident photon (originating from either the laser pulse or ambient light) at \emph{detection time} $\timestamp_\cycle$ after the start of the cycle. It then enters a \emph{dead time}, during which it cannot detect any photons. Once the dead time ends, the SPAD remains inactive until the next pulse emission, at which point the next cycle begins. We note that there can be multiple pulse emissions during a single cycle. After $\numcycles$ cycles, the LiDAR system returns the sequence of detection times $\Timestamps \equiv \curly{\timestamp_1,\dots,\timestamp_\numcycles}$, measured using the sequence of gate times $\Gates \equiv \curly{\gate_1,\dots,\gate_\numcycles}$.
\footnote{Prior works~\cite{Gupta2019AsynchronousS3,Gupta2019PhotonFloodedS3,Heide2018SubpicosecondP3,Rapp2019DeadTC,RappHighFlux} typically study the \emph{\detected}, and not the sequence of detection times. As we discuss in the supplement, the two formulations are consistent.}
We discretize the time between pulse emissions into $\Bins\equiv\floor{\nicefrac{1}{\resolution\ptpfrequency}}$ temporal bins, where the $\bin$-th bin corresponds to the time interval $t\in\left[\bin\cdot\resolution, \paren{\bin+1}\cdot\resolution\right)$ since pulse emission. Then, the range of gate times is $\gate_\cycle \in \curly{0,\dots,\Bins-1}$, and the range of detection times is $\timestamp_\cycle \in \curly{\gate_\cycle,\dots,\gate_\cycle+\Bins-1}$.
\footnote{This assumes that the SPAD always detects a photon during a period of $\nicefrac{1}{\ptpfrequency}$ after it becomes active. In practice, the SPAD may detect no photon. We ignore this for simplicity, and refer to Gupta et al.~\cite{Gupta2019AsynchronousS3} for details.}
We describe a probabilistic model for $\Timestamps$ (Section~\ref{sec:probability}), then see how to estimate depth from it (Section~\ref{sec:depth}). 

\vspace{-0.05in}
\subsection{Probabilistic model}\label{sec:probability}
\vspace{-0.05in}
Our model closely follows the asynchronous image formation model of Gupta et al.~\cite{Gupta2019AsynchronousS3}.
\footnote{We refer to the supplement for a discussion of the assumptions made by this model (e.g., infinitesimal pulse duration) and their implications.}
We first consider the \emph{\incident} $\hsincident_\truedepth\bracket{\bin},\, \bin\in\curly{0,\dots,\Bins-1}$: for each bin $\bin$, $\hsincident_\truedepth\bracket{\bin}$ is the number of photons incident on the SPAD during the time interval $t\in\left[\bin\cdot\resolution, \paren{\bin+1}\cdot\resolution\right)$ since the last pulse emission. The subscript $\truedepth$ indicates that the histogram depends on scene depth, as we explain shortly. We model each $\hsincident_\truedepth\bracket{\bin}$ as a Poisson random variable,
\vspace{-.1in}
\begin{equation}
    \label{eq:incident}
    \hsincident_\truedepth\bracket{\bin}\sim\Poisson\paren{\rate_\truedepth\bracket{\bin}},
    \vspace{-.15in}
\end{equation}
with rate equal to,
\vspace{-.15in}
\begin{equation}
    \label{eq:waveform}
    \rate_\truedepth\bracket{\bin} = \bkgflux + \delta_{\bin,\truedepth}\sigflux.
    \vspace{-.15in}
\end{equation}
The function $\rate_\truedepth\bracket{\bin},\, \bin\in\curly{0,\dots,\Bins-1}$ is the \emph{\transient}~\cite{jarabo2014framework,pediredla2019ellipsoidal}. In Equation~\eqref{eq:waveform}, the \emph{ambient flux} $\bkgflux$ is the average number of incident \emph{background} photons (i.e., photons due to ambient light) at the SPAD during time $\resolution$, which we assume to be time-independent. The \emph{signal flux} $\sigflux$ is the average number of incident \emph{signal} photons (i.e., photons due to the laser). $\bkgflux$ and $\sigflux$ depend on scene reflectivity and distance, and the flux of ambient light (for $\bkgflux$) and laser pulses (for $\sigflux$). We refer to their ratio as the \emph{signal-to-background ratio} $\SBR\equiv \nicefrac{\sigflux}{\bkgflux}$.
\footnote{Prior works~\cite{Gupta2019AsynchronousS3,Gupta2019PhotonFloodedS3} define $\SBR\equiv \nicefrac{\sigflux}{\Bins\bkgflux}$. We omit $\Bins$, to make $\SBR$ indicative of the difficulty in resolving the signal bin from the background bins. 
}
$\delta_{i,j}$ is the Kronecker delta, and $\truedepth \equiv \floor{\frac{2z}{c\resolution}}$, where $z$ is the scene distance and $c$ is the speed of light. We use $\truedepth$ as a proxy for depth.

We now consider the $p$-th cycle of the LiDAR operation. Given that, the SPAD can only detect the first incident photon after activation, a detection time of $\timestamp_\cycle$ means that: i) there were no incident photons during the time bins $\curly{\gate_\cycle, \timestamp_\cycle-1}$; and ii) there was at least one incident photon at time bin $\timestamp_\cycle$. The probability of this event occurring is:
\footnote{We use the notational convention that $\Timestamp_\cycle$, $\Gate_\cycle$, $\Depth$ are random variables, and $\timestamp_\cycle$, $\gate_\cycle$, $\truedepth$ are specific values for these random variables.}
\vspace{-.15in}
\begin{align}
    &\Prob\curly{\Timestamp_\cycle=\timestamp_\cycle \mid \Gate_\cycle=\gate_\cycle, \Depth=\truedepth} = \nonumber \\
    &\Prob\curly{\hsincident_\truedepth\bracket{\timestamp_\cycle \bmod \Bins} \ge 1} \!\!\prod_{\timestamp=\gate_\cycle}^{\timestamp_\cycle-1}\!\!\Prob\curly{\hsincident_\truedepth\bracket{\timestamp \bmod \Bins} = 0}.
    \label{eq:timestamp_prob}
\end{align}
To simplify notation, in the rest of the paper we use: \vspace{-.1in}
\begin{equation}
    \vspace{-.15in}
    \label{eq:tslikelihood}
    \tslikelihood{\timestamp}{\gate}{\truedepth}\equiv\Prob\curly{\Timestamp=\timestamp \mid \Gate=\gate, \Depth=\truedepth}.
\end{equation}
Using Equation~\eqref{eq:incident}, we can rewrite this probability as: \vspace{-.15in}
\begin{equation}
    \!\!\tslikelihood{\timestamp_\cycle}{\gate_\cycle}{\truedepth}\!=\! \paren{1-e^{-\rate_\truedepth\bracket{\timestamp_\cycle \bmod \Bins}}}\!\!\prod_{\timestamp=\gate_\cycle}^{\timestamp_\cycle-1}\!\!e^{-\rate_\truedepth\bracket{\timestamp \bmod \Bins}}.
    \vspace{-.15in}
    \label{eq:timestamp_pdf}
\end{equation}
Lastly, we define the \emph{detection sequence likelihood}: \vspace{-.1in}
\begin{align}
    \nonumber
    \tslikelihood{\Timestamps}{\Gates}{\truedepth} \equiv \Prob\{&\Timestamp_1=\timestamp_1,\dots,\Timestamp_\numcycles = \timestamp_\numcycles \mid \\
    \label{eq:measlikelihood}
    &\Gate_1=\gate_1,\dots,\Gate_\numcycles = \gate_\numcycles, \Depth=\truedepth\}.
\end{align}
Given that the detection times are conditionally independent of each other given the gate times, we have
\vspace{-.15in}
\begin{equation}
	\vspace{-.1in}
    \label{eq:measlikelihood:product}
    \tslikelihood{\Timestamps}{\Gates}{\truedepth} = \prod_{\cycle=1}^{\numcycles} \tslikelihood{\timestamp_\cycle}{\gate_\cycle}{\truedepth}.
\end{equation}
Equations~\eqref{eq:waveform},~\eqref{eq:timestamp_pdf}, and~\eqref{eq:measlikelihood:product} fully determine the probability of a sequence of detection times $\Timestamps$, measured using a sequence of gate times $\Gates$, assuming scene depth $\truedepth$.

\boldstart{Pile-up.} We consider the case where we fix $\gate_\cycle=0$ for all cycles $\cycle=1,\dots,\numcycles$; that is, gating always activates the SPAD at the start of a cycle. Then, Equation~\eqref{eq:timestamp_pdf} becomes
\vspace{-.1in}
\begin{equation}
	\vspace{-.1in}
    \label{eq:pileup}
    \tslikelihood{\bin}{0}{\truedepth}=\paren{1-e^{-\rate_\truedepth\bracket{\bin}}}\prod_{\timestamp=0}^{\bin-1}e^{-\rate_\truedepth\bracket{\timestamp}},\, \bin\in{0,\dots,\Bins}.
\end{equation}
Equations~\eqref{eq:pileup} and~\eqref{eq:waveform} show that, when the ambient flux $\bkgflux$ is large (e.g., outdoors operation), the probability of detecting a photon at a later time bin $\bin$ is small. This effect, termed \emph{pile-up}~\cite{Coates1968TheCF,Gupta2019PhotonFloodedS3}, can result in inaccurate depth estimates as scene depth increases. As we discuss in Section~\ref{sec:gating}, carefully selected gate sequences $\Gates$ can mitigate pile-up. 

\vspace{-0.075in}
\subsection{Depth estimation}\label{sec:depth}
\vspace{-0.05in}
We now describe how to use the probabilistic model of Section~\ref{sec:probability} to estimate the scene depth $\truedepth$ from $\Timestamps$ and  $\Gates$.

\boldstart{Coates' depth estimator.} Gupta et al.~\cite{Gupta2019PhotonFloodedS3,Gupta2019AsynchronousS3} adopt a two-step procedure for estimating depth. First, they form the maximum likelihood (ML) estimate of the \transient\ given the detection and gate sequences:
\vspace{-.1in}
\begin{equation}
   	\vspace{-.1in}
    \label{eq:mle}
    \curly{\hat{\rate}\bracket{\bin}}_{\bin=0}^{\Bins-1} \equiv \argmax_{\curly{\rate\bracket{\bin}}_{\bin=0}^{\Bins-1}} \Prob\curly{\Timestamps \mid \Gates, \curly{\rate\bracket{\bin}}_{\bin=0}^{\Bins-1}}.
\end{equation}
The likelihood function in Equation~\eqref{eq:mle} is analogous to that in Equations~\eqref{eq:timestamp_pdf} and~\eqref{eq:measlikelihood:product}, with an important difference: Whereas Equation~\eqref{eq:timestamp_pdf} assumes that the \transient\ has the form of Equation~\eqref{eq:waveform}, the ML problem of Equation~\eqref{eq:mle} makes no such assumption and estimates an arbitrarily-shaped \transient\ $\rate\bracket{\bin},\,\bin\in\curly{0,\dots\Bins-1}$. Gupta et al.~\cite{Gupta2019AsynchronousS3} derive a closed-form expression for the solution of Equation~\eqref{eq:mle}, which generalizes the \emph{Coates' estimate} of the \transient~\cite{Coates1968TheCF} for arbitrary gate sequences $\Gates$.

Second, they estimate depth as:
\vspace{-.1in}
\begin{equation}
    \label{eq:coates}
    \coatesdepth\paren{\Timestamps,\Gates} \equiv \argmax_{\bin\in 0,\dots,\Bins-1} \hat{\rate}\bracket{\bin}.
    \vspace{-.1in}
\end{equation}
This estimate assumes that the true underlying \transient\ is well-approximated by the $\rate_\truedepth$ model of Equation~\eqref{eq:waveform}. We refer to $\coatesdepth\paren{\Timestamps,\Gates}$ as the \emph{Coates' depth estimator}.



\boldstart{MAP depth estimator.} If we assume that the \transient\ has the form $\rate_\truedepth$ of Equation~\eqref{eq:waveform}, then Equations~\eqref{eq:timestamp_pdf} and~\eqref{eq:measlikelihood:product} directly connect the detection times $\Timestamps$ and depth $\truedepth$, eschewing the \transient. If we have available some prior probability $\prior\paren{\truedepth},\, \truedepth\in\curly{0,\dots,\Bins-1}$ on depth, we can use Bayes' rule to compute the \emph{depth posterior}:
\vspace{-.1in}
\begin{equation}
	\vspace{-.1in}
    \label{eq:depthposterior}
    \tslikelihood{\truedepth}{\Gates}{\Timestamps} \equiv \frac{\tslikelihood{\Timestamps}{\Gates}{\truedepth}\prior\paren{\truedepth}}{\sum_{\truedepth^\prime=0}^{\Bins-1}\tslikelihood{\Timestamps}{\Gates}{\truedepth^\prime}\prior\paren{\truedepth^\prime}}.
\end{equation}
We adopt maximum a-posteriori (MAP) estimation:
\vspace{-.1in}
\begin{equation}
	\vspace{-.1in}
    \label{eq:mapdepth}
    \mapdepth\paren{\Timestamps,\Gates} \equiv \argmax_{\truedepth\in \curly{0,\dots,\Bins-1}} \tslikelihood{\truedepth}{\Gates}{\Timestamps}.
\end{equation}
When using a uniform prior, the depth posterior $\tslikelihood{\truedepth}{\Gates}{\Timestamps}$ and the detection sequence likelihood $\tslikelihood{\Timestamps}{\Gates}{\truedepth}$ are equal, and the \emph{MAP depth estimator} $\mapdepth\paren{\Timestamps,\Gates}$ is also the ML depth estimator. We note that Heide et al.~\cite{Heide2018SubpicosecondP3} proposed a similar MAP estimation approach, using a total variation prior that jointly constrains depth at nearby scan points.

It is worth comparing the MAP $\mapdepth\paren{\Timestamps,\Gates}$ and Coates' $\coatesdepth\paren{\Timestamps,\Gates}$ depth estimators. First, the two estimators have similar computational complexity. This is unsurprising, as the expressions for the depth posterior of Equation~\eqref{eq:depthposterior} and the Coates' estimate of Equation~\eqref{eq:mle} are similar, both using the likelihood functions of Equations~\eqref{eq:timestamp_pdf} and~\eqref{eq:measlikelihood:product}. A downside of the MAP estimator is that it requires knowing $\bkgflux$ and $\sigflux$ in Equation~\eqref{eq:waveform}. In practice, we found it sufficient to estimate the background flux $\bkgflux$ using a small percentage $\paren{\approx 2\%}$ of the total SPAD cycles, and to marginalize the signal flux $\sigflux$ using a uniform prior.

\begin{figure}[t]
  \centering
   \includegraphics[width=\linewidth]{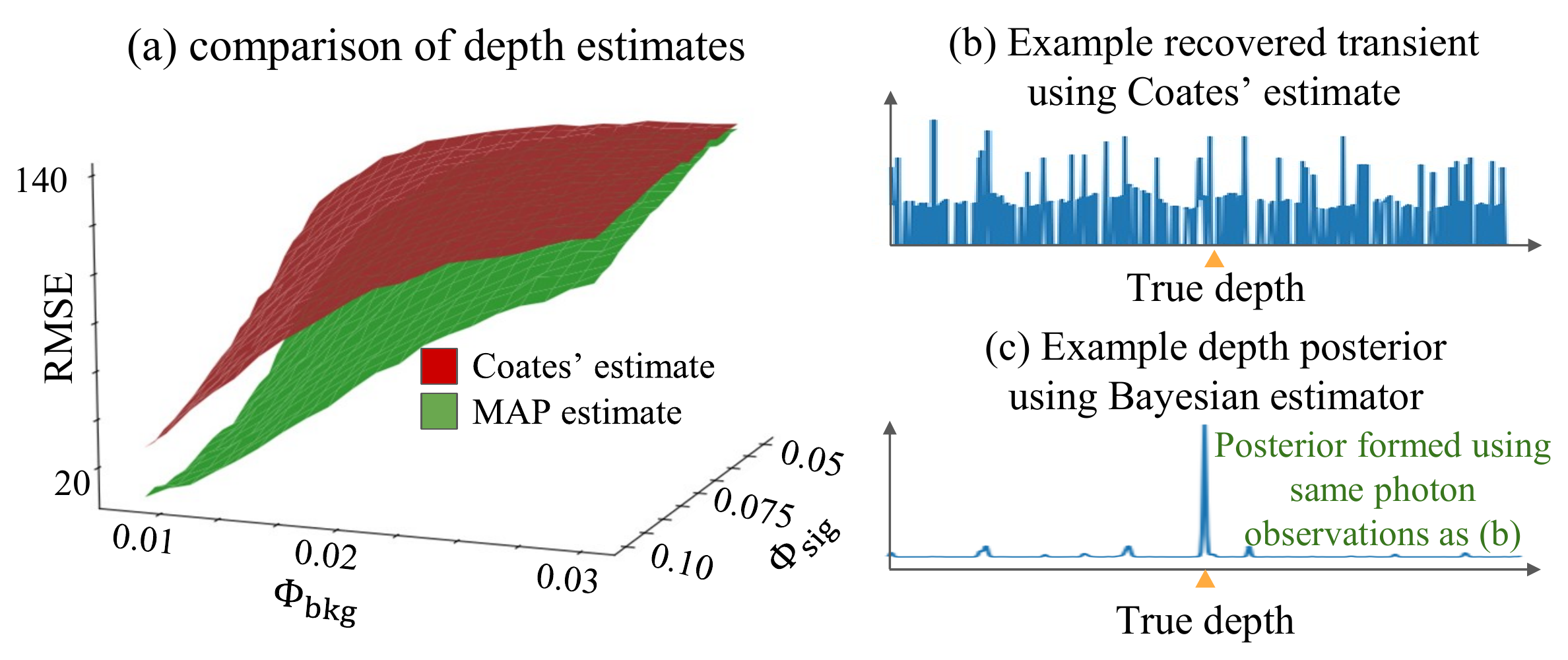}
   \vspace{-0.3in}
   \caption{\textbf{Relative performance of MAP estimator and Coates' estimator for depth recovery.} (a) Bayesian estimator consistently outperforms Coates' estimator across different ambient and signal flux levels. (b) Example recovered transient using Coates' estimator illustrates case where bins with high variance estimates may be mistaken for true depth. (c) Depth posterior formed using the same photon observations as (b) shows a distinct peak at true depth.}
   \label{fig:posterior_coates}
   \vspace{-0.25in}
\end{figure}

Second, when using a uniform prior, the MAP estimator can provide more accurate estimates than the Coates' estimator, in situations where the \transient\ model of Equation~\eqref{eq:waveform} is accurate. To quantify this advantage, we applied both estimators on measurements we simulated for different values of background and signal flux, assuming a uniform gating scheme~\cite{Gupta2019AsynchronousS3}: As we see in Figure~\ref{fig:posterior_coates}, the MAP estimator outperforms the Coates' estimator, especially when ambient flux is significantly higher than signal flux. By contrast, the Coates' estimator can be more accurate than the MAP estimator when the \transient\ deviates significantly from Equation~\eqref{eq:waveform}. This can happen due to multiple peaks (e.g., due to transparent or partial occluders) or indirect illumination (e.g., subsurface scattering, interreflections). In the supplement, we show that, when we combine the MAP estimator with our adaptive gating scheme of Section~\ref{sec:gating}, we obtain correct depth estimates even in cases of such model mismatch. 

Third, the MAP estimator allows incorporating, through the prior, available side information about depth (e.g., from scans at nearby pixels, or from an RGB image~\cite{Xia2020GeneratingAE}). Before we conclude this section, we mention that the MAP estimator is the \emph{Bayesian estimator} with respect to the $\loss$ loss~\cite{berger2013statistical}:
\vspace{-.25in}
\begin{equation}
	\vspace{-.1in}
    \label{eq:bayesian}
    \mapdepth\paren{\Timestamps,\Gates} = \argmin_{\truedepth\in \curly{0,\dots,\Bins-1}} \Exp{\truedepth^\prime\sim\tslikelihood{\truedepth^\prime}{\Gates}{\Timestamps}}{\loss\paren{\truedepth, \truedepth^\prime}},
\end{equation}
\vspace{-.in}
where
\begin{equation}
    \loss\paren{x,y} \equiv \begin{cases}
    0, &\mbox{if}\; x = y. \\
    1, &\mbox{otherwise}.
    \end{cases}
    \vspace{-.1in}
\end{equation}
We will use this fact in the next section, as we use the depth posterior and MAP estimator to develop adaptive gating.

\begin{figure*}
  	  \centering
	  \includegraphics[width=\textwidth]{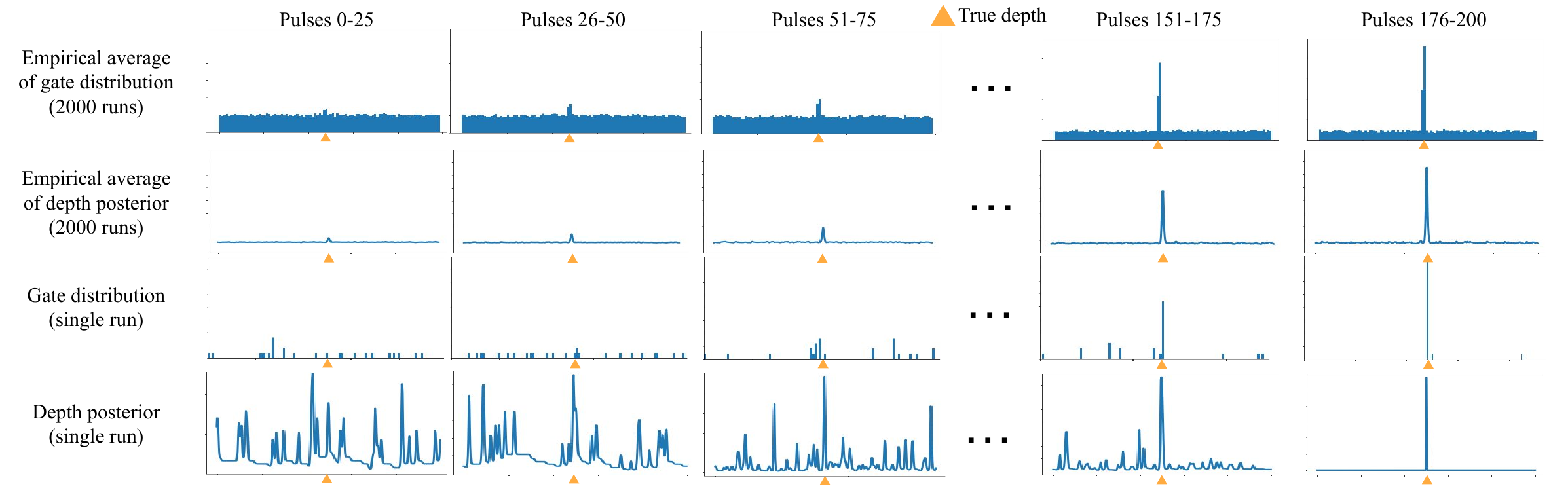}
      \caption{\textbf{Evolution of gate selection using adaptive gating.} Adaptive gating samples gate location based on the depth posterior formed using previous photon arrivals. Initial gates are approximately uniformly distributed, since depth posterior formed under lower number of photon observations have high variance. As more photons are observed, the depth posterior begins to form a sharper peak around true depth, causing gates selected through Thompson sampling to converge.}
  \label{fig:gates}
     \vspace{-0.25in}
\end{figure*}

\vspace{-0.1in}
\section{Adaptive Gating}\label{sec:gating}
\vspace{-0.1in}
We now turn our attention to the selection of the gate sequence $\Gates$. As we mentioned in Section~\ref{sec:background}, we aim to use gating to mitigate pile-up. We briefly review two prior gating schemes, then introduce our \emph{adaptive gating}.

\boldstart{Fixed gating.} A fixed gating scheme uses the same gate for all $\numcycles$ TCSPC cycles, $\gate_\cycle = \gate_\text{fixed},\, \cycle\in\curly{1,\dots,\numcycles}$. So long as this fixed gate is before the scene depth, $\gate_\text{fixed} < \truedepth$, it will prevent the detection of early-arriving photons due to ambient light, and thus increase the probability of detection of signal photons. For fixed gating to be effective, $\gate_\text{fixed}$ should be close, and ideally \emph{equal} to the true depth $\truedepth$, as setting s$\gate_\text{fixed}=\truedepth$ maximizes the detection probability $\tslikelihood{\truedepth}{\gate_\text{fixed}}{\truedepth}$ in Equation~\eqref{eq:timestamp_pdf}. Unfortunately, this requires knowing the true depth $\truedepth$, or at least a reliable estimate thereof; such an estimate is generally available only after several cycles.

\boldstart{Uniform gating.} Gupta et al.~\cite{Gupta2019AsynchronousS3} introduced a uniform gating scheme, which distributes gates uniformly across the entire depth range. If, for simplicity, we assume that the numbers of cycles and temporal bins are equal, $\numcycles=\Bins$, then uniform gating sets $\gate_\cycle = \cycle,\, \cycle\in\curly{1,\dots,\numcycles}$. 
This maximizes the detection probability of each bin for a few cycles, and ``averages out'' pile-up effects. Compared to fixed gating, uniform gating does not require an estimate of the true depth $\truedepth$. Conversely, uniform gating cannot take advantage of increasing information about $\truedepth$ as more cycles finish.

Gupta et al.~\cite{Gupta2019AsynchronousS3} propose using the SPAD in free-running mode without gating---the SPAD becomes active immediately after dead time ends---as an alternative to uniform gating. As they explain, using free-running mode also ensures that all bins have high probability of detection for a few cycles, similar to uniform gating; and provides additional advantages (e.g., maximizes SPAD active time, simplifies hardware). Therefore, we often compare against free-running mode instead of uniform gating. 


\boldstart{Desired behavior for adaptive gating.} Before formally describing our adaptive gating scheme, we describe at a high-level the desired behavior for such a scheme. Intuitively, an ideal gating scheme should behave as a hybrid between fixed and uniform gating. During the early stages of LiDAR operation (first few cycles) we have little to no information about scene depth---all temporal bins have approximately equal probability of being the true depth. Thus, a hybrid scheme should mimic uniform gating to explore the entire depth range. During the later stages of LiDAR operation (last few cycles), we have rich information about scene depth from the detection times recorded during preceding cycles---only one or few temporal bins have high probability of being the true depth. Thus, a hybrid scheme should mimic (near-fixed) gating, to maximize the detection probability of the few remaining candidate temporal bins. At intermediate stages of LiDAR operation, the hybrid scheme should progressively transition from uniform towards fixed gating, with this progression adapting from cycle to cycle to the information about scene depth available from previously-recorded detection times.


\boldstart{Thompson sampling.} To turn the above high-level specification into a formal algorithm, we use two building blocks. First, we use the probabilistic model of Section~\ref{sec:background} to quantify the information we have about scene depth $\truedepth$ at each cycle. At the start of cycle $\cycle$, the LiDAR has recorded detection times $\Timestamps_{\cycle-1} \equiv \curly{\timestamp_q, q=1,\dots,\cycle}$ using gate times $\Gates_{\cycle-1} \equiv \curly{\gate_q, q=1,\dots,\cycle}$. Then, the depth posterior $\tslikelihood{\truedepth}{\Gates_{\cycle-1}}{\Timestamps_{\cycle-1}}$ of Equation~\eqref{eq:depthposterior} represents all the information we have available about scene depth, from both recorded detection times and any prior information ($\prior\paren{\truedepth}$). 

Second, we use \emph{Thompson sampling}~\cite{thompson1933likelihood} to select the gate times $\gate_\cycle$. Thompson sampling is a classical algorithm for online experimental design: This is the problem setting of deciding on the fly parameters of a sequence of experiments, using at any given time available information from all experiments up to that time, in a way that maximizes some utility function~\cite{russo2017tutorial}. Translating this into the context of SPAD-based LiDAR, the sequence of experiments is the $\numcycles$ TCSPC cycles; at the $\cycle$-th experiment, the parameter to be decided is the gate time $\gate_\cycle$. and the available information is the depth posterior $\tslikelihood{\truedepth}{\Gates_{\cycle-1}}{\Timestamps_{\cycle-1}}$; lastly the utility function is the accuracy of the final depth estimate. Thompson sampling selects each gate $\gate_\cycle$ by first sampling a \emph{depth hypothesis} $\hypotdepth$ from the depth posterior $\tslikelihood{\truedepth}{\Gates_{\cycle-1}}{\Timestamps_{\cycle-1}}$, and then finding the gate time that maximizes a \emph{reward function} $\reward\paren{\hypotdepth, g}$. Algorithm~\ref{alg:gating} shows the resulting adaptive gating scheme (blue lines correspond to modifications we describe in Section~\ref{sec:exposure}).
%
\begin{algorithm}[t]
    \caption{\label{alg:gating} Adaptive gating \color{blue}{with adaptive exposure}.}
    \SetAlgoLined
    \SetCommentSty{mycmtfn}
    \SetKwComment{tccinline}{// }{}
    \KwIn{max number of cycles $\numcycles$, depth prior $\prior\paren{\truedepth}$.}
    \KwOut{depth estimate $\mapdepth\paren{\Timestamps_\cycle,\Gates_\cycle}$}
    \tcc{Initialization}
    $\cycle \gets 0$ \tccinline*{initialize cycle counter}
    $\tslikelihood{\truedepth}{\Gates_0}{\Timestamps_0}\gets \prior\paren{\truedepth}$ \tccinline*{initialize depth posterior}
    \tcc{Acquisition}
    \While{$\cycle \leq \numcycles$}{
        $\cycle\!\gets\!\cycle + 1$ \tccinline*{start next cycle}
        $\hypotdepth\!\sim\!\tslikelihood{\truedepth}{\Gates_{\cycle-1}}{\Timestamps_{\cycle-1}}$ \tccinline*{sample depth hypothesis}
        $\gate_\cycle\!\gets\!\hypotdepth$ \tccinline*{select gate (Proposition~\ref{prop:gate})} 
        $\timestamp_\cycle\!\gets\!\text{TCSPC}\paren{\gate_\cycle}$ \tccinline*{record detection time}
        $\tslikelihood{\truedepth}{\Gates_{\cycle}}{\Timestamps_{\cycle}}\!\propto\!p^{\gate_{\cycle}}_{\truedepth}\!\paren{\timestamp_\cycle} p^{\Gates_{\cycle-1}}_{\Timestamps_{\cycle-1}}\!\paren{\truedepth}$ \tccinline*{update depth posterior}
        \color{blue}
        $\mapdepth\paren{\Timestamps_\cycle,\Gates_\cycle}\!\gets\!\argmax_\truedepth \tslikelihood{\truedepth}{\Gates_\cycle}{\Timestamps_\cycle}$ \tccinline*{update depth estimate}
        $\termination\paren{\Timestamps_\cycle,\Gates_\cycle}\!\gets\!1\!-\!\tslikelihood{\mapdepth\paren{\Timestamps_\cycle,\Gates_\cycle}}{\Gates_\cycle}{\Timestamps_\cycle}$ \tccinline*{compute termination function (Equation~\eqref{eq:termination})}
        \If{$\termination\paren{\Timestamps_\cycle,\Gates_\cycle} < \epsilon$}{
            \algobreak; \tccinline*{terminate acquisition}
        }
        \color{black}
    }
    \tcc{Final depth estimation}
    $\mapdepth\paren{\Timestamps_\cycle,\Gates_\cycle}\gets\argmax_\truedepth \tslikelihood{\truedepth}{\Gates_\cycle}{\Timestamps_\cycle}$;
\end{algorithm}
%
%
\boldstart{Reward function.} We motivate our choice of reward function as follows: At cycle $\cycle$, Thompson sampling \emph{assumes} that the true depth equals the depth hypothesis $\hypotdepth$ sampled from the depth posterior. Equivalently, Thompson sampling assumes that the detection time we will measure after cycle $\cycle$ concludes is distributed as $\timestamp_\cycle\sim\tslikelihood{\timestamp}{\gate_\cycle}{\hypotdepth}$ (Equation~\eqref{eq:timestamp_pdf}). As we aim to infer depth, we should select a gate $\gate_\cycle$ such that, if we estimated depth from only the next detection $\timestamp_\cycle$, the resulting estimate would be \emph{in expectation} close to the depth hypothesis $\hypotdepth$ we assume to be true. Formally,
\begin{equation}
    \label{eq:reward}
    \reward\paren{\hypotdepth, \gate} \equiv -\Exp{\timestamp\sim\tslikelihood{\timestamp}{\gate}{\hypotdepth}}{\loss\paren{\mapdepth\paren{\timestamp,\gate}, \hypotdepth}}.
\end{equation}
In Equation~\eqref{eq:reward}, to estimate depth from the expected detection time $\timestamp$, we use the same MAP depth estimator of Equation~\eqref{eq:mapdepth} as for the final depth estimate $\mapdepth\paren{\Timestamps,\Gates}$. The MAP depth estimator is optimal with respect to the $\loss$ loss (Equation~\eqref{eq:bayesian}), thus we use the same loss for the reward function. 
Selecting the gate $\gate_\cycle$ requires maximizing the reward function $\reward\paren{\hypotdepth, \gate}$, which we can do analytically.

\begin{prop} 
\label{prop:gate}
The solution to the optimization problem
\vspace{-.075in}
\begin{equation}
    \tilde{\gate} \equiv \argmax_{\gate\in\curly{0,\dots,\Bins-1}} \reward\paren{\hypotdepth, g}
    \vspace{-.125in}
\end{equation}
for the reward function of Equation~\eqref{eq:reward} equals $\tilde{\gate} = \hypotdepth$.
\vspace{-.1in}
\end{prop}
We provide the proof in the supplement. Intuitively, minimizing the expected $\loss$ loss between the estimate $\mapdepth\paren{\timestamp,\gate}$ and the depth hypothesis $\hypotdepth$ is equivalent to maximizing the probability that $\timestamp \bmod \Bins = \hypotdepth$; that is, we want to maximize the probability that a photon detection occurs at the same temporal bin as the depth hypothesis. We do this by setting the gate equal to the depth hypothesis, $\gate_\cycle=\hypotdepth$. 
\footnote{In practice, we set the gate time $\gate_\cycle$ a few bins before the depth hypothesis $\hypotdepth$, to account for the finite laser pulse width and timing jitter.}

\boldstart{Intuition behind Thompson sampling.} Before concluding this section, we provide some intuition about how Thompson sampling works, and why it is suitable for adaptive gating. We can consider adaptive gating with Thompson sampling as a procedure for balancing the exploration-exploitation trade-off. Revisiting the discussion at the start of this section, fixed gating maximizes \emph{exploitation}, by only gating at one temporal bin (or a small number thereof); conversely, uniform gating maximizes \emph{exploration}, by gating uniformly across the entire depth range. During the first few cycles, adaptive gating maximizes exploration: as only few measurements are available, the depth posterior is flat, and depth hypotheses (and thus gate times) are sampled approximately uniformly as with uniform gating. As the number of cycles progresses, adaptive gating shifts from exploration to exploitation: additional measurements make the depth posterior concentrated around a few depth values, and gate times are sampled mostly among those. After a sufficiently large number of cycles, adaptive gating maximizes exploitation: the depth posterior peaks at a single depth, and gate times are almost always set to that depth, as in fixed gating. Figure~\ref{fig:gates} uses simulations to visualize this transition from exploration to exploitation. This behavior matches the one we set out to achieve at the start of this section. Lastly, we mention that Thompson sampling has strong theoretical guarantees for \emph{asymptotic} optimality~\cite{russo2017tutorial}; this suggests that our adaptive gating scheme balances the exploration-exploitation trade-off in a way that, asymptotically (given enough cycles), maximizes depth accuracy.

\vspace{-.1in}
\section{Adaptive Exposure}\label{sec:exposure}
\vspace{-.1in}

The accuracy of a depth estimate from SPAD measurements depends on three main factors: the exposure time (i.e., number of laser pulses, which also affects the number of cycles $\numcycles$), ambient flux $\bkgflux$, and signal-to-background ratio $\SBR$. For a fixed exposure time, increasing ambient flux or lowering SBR will result in higher depth estimation uncertainty. Even under conditions of identical ambient flux and SBR, the required exposure time to reach some uncertainty threshold can vary significantly, because of the random nature of photon arrivals and detections. 

In a generic scene, different pixels can have very different ambient flux or SBR (e.g., due to cast shadows, varying reflectance, and varying depth). Therefore, using a fixed exposure time will result in either a lot of wasted exposure time on pixels for which shorter exposure times are sufficient, or high estimation uncertainty in pixels that require a longer exposure time. Ideally, we want to \emph{adaptively} extend or shorten the per-pixel exposure time, depending on how many TCSPC cycles are needed to reach some desired depth estimation uncertainty threshold.

\boldstart{Adaptive gating with adaptive exposure.} The adaptive gating scheme of Section~\ref{sec:gating} lends itself to a modification that also adapts the number of cycles $\numcycles$, and thus exposure time. In Algorithm~\ref{alg:gating}, we can terminate the while loop early at a cycle $\cycle \le \numcycles$, if the depth posterior  $\tslikelihood{\truedepth}{\Gates_{\cycle}}{\Timestamps_{\cycle}}$ becomes concentrated enough that we expect depth estimation to have low error. Formally, we define a \emph{termination function} as the expected error with respect to the depth posterior:
\vspace{-.15in}
\begin{align}
  \vspace{-.15in}
  \label{eq:termination}
  \termination\paren{\Timestamps_\cycle,\Gates_\cycle} &\equiv \Exp{\truedepth\sim\tslikelihood{\truedepth}{\Gates_\cycle}{\Timestamps_\cycle}}{\loss\paren{\mapdepth\paren{\Timestamps_\cycle,\Gates_\cycle}, \truedepth}} \\\vspace{-.15in}
  &= 1 - \tslikelihood{\mapdepth\paren{\Timestamps_\cycle,\Gates_\cycle}}{\Gates_\cycle}{\Timestamps_\cycle}.
\end{align}
In Equation~\eqref{eq:termination}, we use the same MAP depth estimator as for the final depth estimate $\mapdepth\paren{\Timestamps,\Gates}$, and the $\loss$ loss for which the MAP estimator is optimal (Equation~\eqref{eq:bayesian}). These choices are analogous to our choices for the definition of the reward function $\reward\paren{\hypotdepth, \gate}$ in Equation~\eqref{eq:reward}. At the end of each cycle $\cycle$, we check whether the termination function $\termination\paren{\Timestamps_\cycle,\Gates_\cycle}$ is smaller than some threshold, and terminate acquisition if it is true. In Algorithm~\ref{alg:gating}, we show in blue how we modify our original adaptive gating algorithm to include this adaptive exposure procedure.

\begin{figure}[t]
	\centering
	\includegraphics[width=1\linewidth]{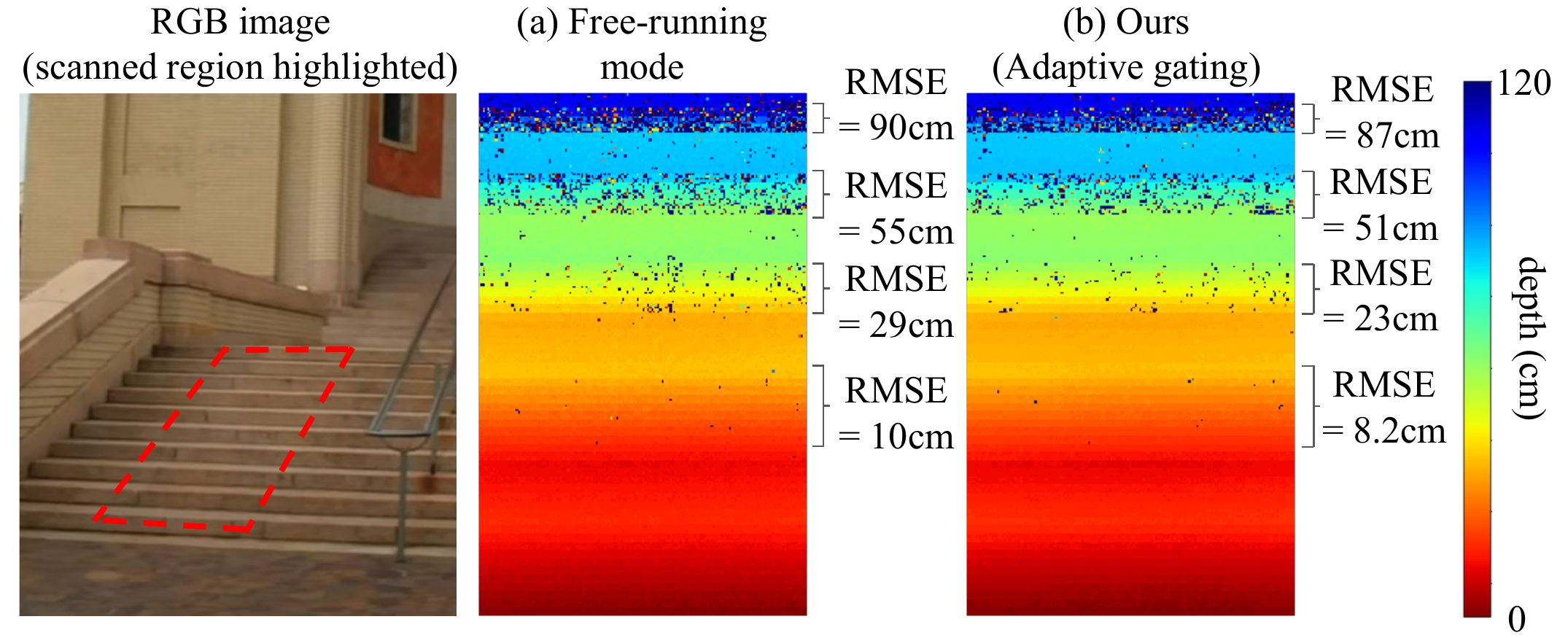}
    \vspace{-0.3in}
	\caption{\textbf{Effect of SBR.} 
		The ``Stairs" scene has regions with significant variations in SBR. 
		Far stairs have lower SBR than nearby stairs due to inverse square falloff. 
		Our adaptive gating scheme achieves lower RMSE than the free-running mode at all the SBRs.}
	\vspace{-0.27in}
	\label{fig:stairs}
\end{figure}

\begin{figure*}
  \centering
   \includegraphics[width=1\linewidth]{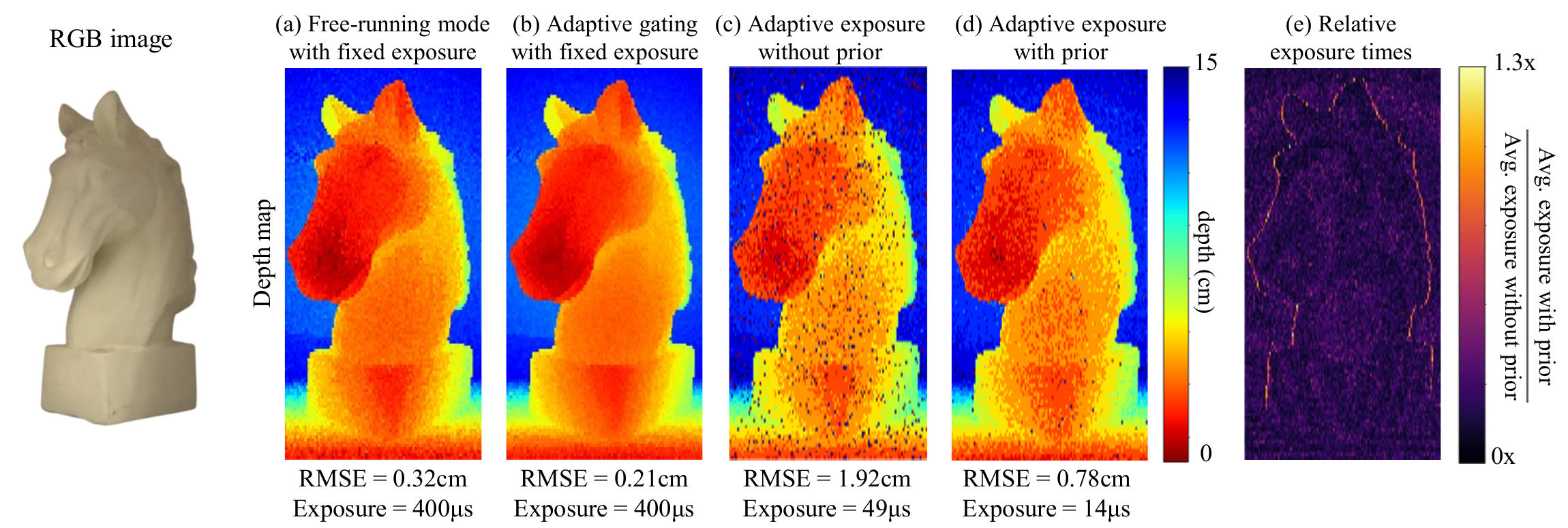}
   \vspace{-0.32in}
   \caption{
	\textbf{3D scanning under indoor illumination.} A horse bust under ambient light of $\phi_\text{bkg} \approx 0.01$.
	The scene does not result in a substantial pile-up like the outdoor scenes.
	Adaptive gating still shows better depth reconstruction than the free-running mode.
   }
	\vspace{-0.18in}
   \label{fig:horse}
\end{figure*}

\begin{figure*}
  \includegraphics[width=\textwidth]{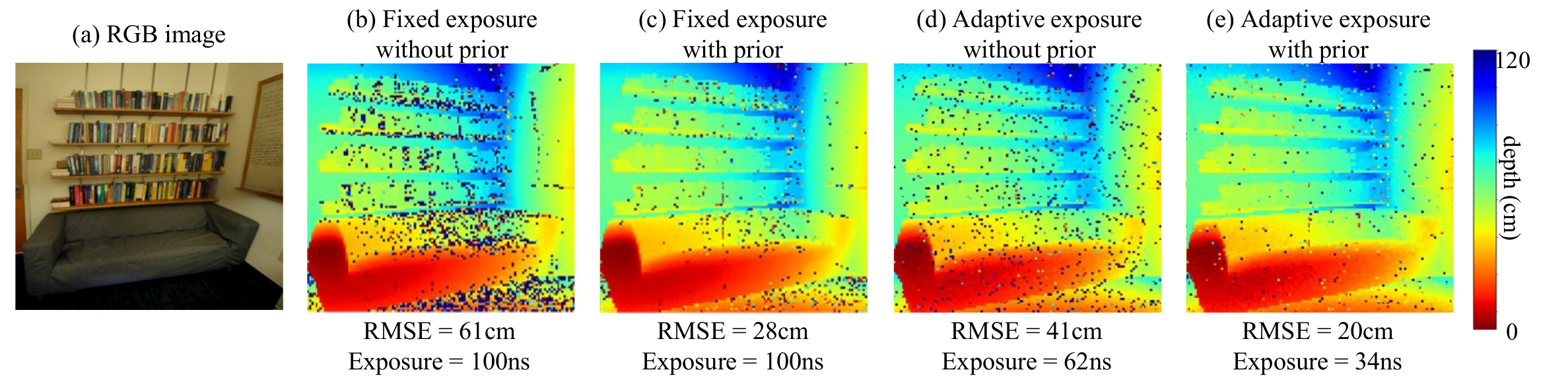}
  \vspace{-0.32in}
  \caption{
	\textbf{Adaptive gating can incorporate external depth priors.}
	We compute a depth prior using neural-network based monocular depth estimation technique \cite{Xia2020GeneratingAE}.
	Incorporating the prior improves both the fixed exposure and adaptive exposure schemes. 
  }
	\label{fig:office}
 \vspace{-0.25in}
\end{figure*}

\vspace{-.1in}
\section{Experimental Results}\label{sec:experiments}
\vspace{-.1in}

We show comparisons using real data from an experimental prototype in the paper, and additional comparisons on real and simulated data in the supplement. Throughout the paper, we use root mean square error (RMSE) as the performance metric, as is common in prior work. In the supplement, we additionally report $loss$ error metrics; this makes performance improvements more pronounced, as our technique optimizes $loss$ error (Section~\ref{sec:gating}). Our code and data are available on the project website~\cite{ProjectWebsite}.

\boldstart{Prototype.} Our SPAD-based LiDAR prototype comprises a fast-gated SPAD (Micro-photon Devices), a $\unit[532]{nm}$ picosecond pulsed laser (NKT Photonics NP100-201-010), a TCSPC module (PicoHarp 300), and a programmable picosecond delayer (Micro-photon Devices). We operated the SPAD in triggered mode (adaptive gating) or free-running mode, with the programmable dead time set to $\unit[81]{ns}$. 
We set the laser pulse frequency to $\unit[10]{MHz}$, with measurements discretized to $500$ bins ($\unit[100]{ps}$ resolution). The raster scan resolution is $128\times 128$, and acquisition time per scan point is $\unit[100]{\mu s}$, for a total acquisition time of $\unit[1.6]{s}$. We note that, if gate selection happens during dead time---easily achievable with optimized compute hardware---our procedure does not introduce additional acquisition latency.

\boldstart{Outdoor scenes.} Outdoor experiments (Figures~\ref{fig:teaser} and~\ref{fig:stairs}) were under direct sunlight (no shadows or clouds) around noon (11 am to 2 pm in the United States), with an estimated background strength of 0.016 photons per pulse per bin. Figure~\ref{fig:teaser} shows a set of 3D reconstructions for the ``Leaf'' scene, captured at noon with a clear sky and under direct sunlight. 
Figures~\ref{fig:teaser}(a)-(b) show depth reconstructions using free-running mode and adaptive gating under fixed exposure times: our method reduces RMSE by around $3\times$. Figure~\ref{fig:teaser}(c) shows the depth reconstruction using adaptive gating combined with adaptive exposure. This combination reduces both RMSE and exposure time (by around $3\times$) compared to free-running mode. Figure~\ref{fig:stairs} shows depth reconstructions for the ``Stairs'' scene, captured under the same sunlight conditions. This scene has significant SBR variations, and we observe that adaptive gating outperforms free-running mode in all SBR regimes.

\boldstart{Indoor scenes.} Figure~\ref{fig:horse} shows reconstructions of the ``Horse'' scene, an indoor scene of a horse bust placed in a light booth. 
Under fixed exposure time, our method achieves $35\%$ lower RMSE compared to free-running mode. We note that RMSE improvement is less pronounced than in outdoor scenes. 
As the indoor scene has significantly higher SBR, this suggests that adaptive gating offers a bigger advantage over free-running mode at low SBRs.

We next examine the effectiveness of employing external priors. In Figure~\ref{fig:horse}(c)-(d), we use a ``flatness'' prior: at each pixel we use a Gaussian prior centered at the depth measured at the previous scanned pixel. Leveraging this simple prior leads to a 70\% decrease in exposure time, and a 60\% decrease in RMSE compared to adaptive gating without the use of a depth prior. We note, however, that this prior is not effective at pixels corresponding to large depth discontinuities. We can see this in Figure~\ref{fig:horse}(e), which visualizes relative exposure time reduction at different pixels. 

In Figure~\ref{fig:office}, we use a monocular depth estimation algorithm~\cite{Xia2020GeneratingAE} to obtain depth estimates and uncertainty values from an RGB image of the ``Office'' scene. We incorporate the monocular prior in our adaptive gating method, and notice a $50\%$ reduction in RMSE for fixed exposure times, and $45\%$ lower total acquisition time using adaptive exposure.
\vspace{-.1in}
\section{Limitations and Conclusion}
\vspace{-.1in}

\boldstart{Active time.} As Gupta et al.~\cite{Gupta2019AsynchronousS3} explain, using free-running mode provides similar advantages as uniform gating, and at the same time maximizes the time during which the SPAD is active---and thus maximizes photon detections. Adaptive gating, likewise, also results in reduced active time compared to free-running mode. However, experimentally we found that, despite the reduced photon detections, our adaptive gating scheme still results in improved depth accuracy for all practical dead time values (see Section~\ref{sec:experiments} and additional experiments evaluating the effect of active and dead time in the supplement). 

\boldstart{Hardware considerations.} Our adaptive gating scheme requires using a SPAD with a controllable gate that can be reprogrammed from pulse to pulse. We implemented this using the same gated SPAD hardware as Gupta et al.~\cite{Gupta2019AsynchronousS3} did for their uniform gating scheme. The main additional hardware requirements are electronics that can produce the gate control signals at $\ptpfrequency$ frequency and $\resolution$ resolution. Both gating schemes require considerably more expensive hardware compared to free-running mode operation, which only requires an ungated SPAD. Whether the improved depth accuracy performance justifies the increased hardware cost and complexity is an application-dependent consideration.

Our adaptive exposure scheme additionally requires beam steering hardware that can on the fly change the scan time for any given pixel. Unfortunately, currently this is only possible at the cost of significantly slower overall scanning: Current beam steering solutions for LiDAR must operate in \emph{resonant mode} to enable $\unit[]{kHz}$ scanning rates, which in turn means that the per-pixel scan times are pre-determined by the resonant scan pattern \cite{pittaluga2020towards}. Thus, using adaptive exposure requires operating beam steering at non-resonant mode. This introduces scanning delays that likely outweigh the gains from reduced per-pixel scan times, resulting in an overall slower scanning rate.

Lastly, recent years have seen the emergence of two-dimensional SPAD arrays~\cite{morimoto2020megapixel}. Current prototypes support a shared programmable gate among all pixels. Adaptive gating would require independent per-pixel programmable gates, which can be implemented at increased fabrication cost, and likely decreased sensitive area. As SPAD arrays time-multiplex acquisition across pixels, adaptive exposure does not offer an obvious advantage in this context.

\boldstart{Conclusion.} We introduced an adaptive gating scheme for SPAD-based LiDAR that mitigates the effects of pile-up under strong ambient light conditions. Our scheme uses a Thompson sampling procedure to select a gating sequence that takes advantage of information available from previously-measured laser pulses, to maximize depth estimation accuracy. Our scheme can also adaptively adjust exposure time per-pixel, as necessary to achieve a desired expected depth error. We showed that our scheme can reduce both depth error and exposure time by more than $100\%$ compared to previous SPAD-based LiDAR techniques, including when operating outdoors under strong sunlight.

\boldstart{Acknowledgments.} We thank Akshat Dave, Ankit Raghuram, and Ashok Veeraraghavan for providing the high-power laser for experiments and invaluable assistance in its use; as well as Matthew O'Toole, David Lindell, and Dorian Chan for help with the SPAD sensor. This work was supported by NSF Expeditions award 1730147, NSF CAREER award 2047341, DARPA REVEAL contract HR0011-16-C-0025, and ONR DURIP award N00014-16-1-2906. Ioannis Gkioulekas is supported by a Sloan Research Fellowship.


{\small
\bibliographystyle{ieee_fullname}
\bibliography{adaptive}

\begin{thebibliography}{10}\itemsep=-1pt

\bibitem{Rangwala2020iPhone}
{The iPhone 12 – LiDAR At Your Fingertips. Forbes (12 November 2020)}.
\newblock
  \url{https://www.forbes.com/sites/sabbirrangwala/2020/11/12/the-iphone-12lidar-at-your-fingertips/}.
\newblock Accessed: 2021-11-17.

\bibitem{2018LiDARDF}
Lidar drives forwards.
\newblock {\em Nature Photonics}, 12:441, 2018.

\bibitem{Acconcia2018FastFF}
Giulia Acconcia, A Cominelli, Massimo Ghioni, and Ivan Rech.
\newblock Fast fully-integrated front-end circuit to overcome pile-up limits in
  time-correlated single photon counting with single photon avalanche diodes.
\newblock {\em Optics express}, 26 12:15398--15410, 2018.

\bibitem{anderson2005experimental}
Dean Anderson, Herman Herman, and Alonzo Kelly.
\newblock Experimental characterization of commercial flash ladar devices.
\newblock In {\em International Conference of Sensing and Technology},
  volume~2, pages 17--23. Citeseer, 2005.

\bibitem{barsi2014multi}
Christopher Barsi, Refael Whyte, Ayush Bhandari, Anshuman Das, Achuta Kadambi,
  Adrian~A Dorrington, and Ramesh Raskar.
\newblock Multi-frequency reference-free fluorescence lifetime imaging using a
  time-of-flight camera.
\newblock In {\em Biomedical Optics}, pages BM3A--53. Optical Society of
  America, 2014.

\bibitem{Becker2015AdvancedTS}
Wolfgang Becker.
\newblock Advanced time-correlated single photon counting applications.
\newblock {\em Advanced Time-Correlated Single Photon Counting Applications},
  2015.

\bibitem{berger2013statistical}
James~O Berger.
\newblock {\em Statistical decision theory and Bayesian analysis}.
\newblock Springer Science \& Business Media, 2013.

\bibitem{bergman2020deep}
Alexander~W Bergman, David~B Lindell, and Gordon Wetzstein.
\newblock Deep adaptive lidar: End-to-end optimization of sampling and depth
  completion at low sampling rates.
\newblock In {\em 2020 IEEE International Conference on Computational
  Photography (ICCP)}, pages 1--11. IEEE, 2020.

\bibitem{bhandari2015blind}
Ayush Bhandari, Christopher Barsi, and Ramesh Raskar.
\newblock Blind and reference-free fluorescence lifetime estimation via
  consumer time-of-flight sensors.
\newblock {\em Optica}, 2(11):965--973, 2015.

\bibitem{bruschini2019single}
Claudio Bruschini, Harald Homulle, Ivan~Michel Antolovic, Samuel Burri, and
  Edoardo Charbon.
\newblock Single-photon avalanche diode imagers in biophotonics: review and
  outlook.
\newblock {\em Light: Science \& Applications}, 8(1):1--28, 2019.

\bibitem{Buttafava2015NonlineofsightIU}
Mauro Buttafava, Jessica Zeman, Alberto Tosi, Kevin~W. Eliceiri, and Andreas
  Velten.
\newblock Non-line-of-sight imaging using a time-gated single photon avalanche
  diode.
\newblock {\em Optics express}, 23 16:20997--1011, 2015.

\bibitem{Coates1968TheCF}
P.~B. Coates.
\newblock The correction for photon `pile-up' in the measurement of radiative
  lifetimes.
\newblock {\em Journal of Physics E: Scientific Instruments}, 1:878--879, 1968.

\bibitem{Cominelli2017HighspeedAL}
A Cominelli, Giulia Acconcia, P. Peronio, Massimo Ghioni, and Ivan Rech.
\newblock High-speed and low-distortion solution for time-correlated single
  photon counting measurements: A theoretical analysis.
\newblock {\em The Review of scientific instruments}, 88 12:123701, 2017.

\bibitem{gokturk2004time}
S~Burak Gokturk, Hakan Yalcin, and Cyrus Bamji.
\newblock A time-of-flight depth sensor-system description, issues and
  solutions.
\newblock In {\em 2004 conference on computer vision and pattern recognition
  workshop}, pages 35--35. IEEE, 2004.

\bibitem{Gupta2019AsynchronousS3}
Anant Gupta, Atul Ingle, and Mohit Gupta.
\newblock Asynchronous single-photon 3d imaging.
\newblock {\em 2019 IEEE/CVF International Conference on Computer Vision
  (ICCV)}, pages 7908--7917, 2019.

\bibitem{Gupta2019PhotonFloodedS3}
Anant Gupta, Atul Ingle, Andreas Velten, and Mohit Gupta.
\newblock Photon-flooded single-photon 3d cameras.
\newblock {\em 2019 IEEE/CVF Conference on Computer Vision and Pattern
  Recognition (CVPR)}, pages 6763--6772, 2019.

\bibitem{Heide2018SubpicosecondP3}
Felix Heide, Steven Diamond, David~B. Lindell, and Gordon Wetzstein.
\newblock Sub-picosecond photon-efficient 3d imaging using single-photon
  sensors.
\newblock {\em Scientific Reports}, 8, 2018.

\bibitem{ingle2021passive}
Atul Ingle, Trevor Seets, Mauro Buttafava, Shantanu Gupta, Alberto Tosi, Mohit
  Gupta, and Andreas Velten.
\newblock Passive inter-photon imaging.
\newblock In {\em Proceedings of the IEEE/CVF Conference on Computer Vision and
  Pattern Recognition}, pages 8585--8595, 2021.

\bibitem{Ingle2019HighFP}
Atul Ingle, Andreas Velten, and Mohit Gupta.
\newblock High flux passive imaging with single-photon sensors.
\newblock {\em 2019 IEEE/CVF Conference on Computer Vision and Pattern
  Recognition (CVPR)}, pages 6753--6762, 2019.

\bibitem{Isbaner2016DeadtimeCO}
Sebastian Isbaner, Narain Karedla, Daja Ruhlandt, Simon~Christoph Stein,
  Anna~M. Chizhik, Ingo Gregor, and J{\"o}rg Enderlein.
\newblock Dead-time correction of fluorescence lifetime measurements and
  fluorescence lifetime imaging.
\newblock {\em Optics express}, 24 9:9429--45, 2016.

\bibitem{jarabo2014framework}
Adrian Jarabo, Julio Marco, Adolfo Munoz, Raul Buisan, Wojciech Jarosz, and
  Diego Gutierrez.
\newblock A framework for transient rendering.
\newblock {\em ACM Transactions on Graphics (ToG)}, 33(6):1--10, 2014.

\bibitem{kadambi2016macroscopic}
Achuta Kadambi, Jamie Schiel, and Ramesh Raskar.
\newblock Macroscopic interferometry: Rethinking depth estimation with
  frequency-domain time-of-flight.
\newblock In {\em Proceedings of the IEEE Conference on Computer Vision and
  Pattern Recognition}, pages 893--902, 2016.

\bibitem{lakowicz1992fluorescence}
Joseph~R Lakowicz, Henryk Szmacinski, Kazimierz Nowaczyk, Klaus~W Berndt, and
  Michael Johnson.
\newblock Fluorescence lifetime imaging.
\newblock {\em Analytical biochemistry}, 202(2):316--330, 1992.

\bibitem{lee2019coding}
Jongho Lee, Jenu~Varghese Chacko, Bing Dai, Syed~Azer Reza, Abdul~Kader Sagar,
  Kevin~W Eliceiri, Andreas Velten, and Mohit Gupta.
\newblock Coding scheme optimization for fast fluorescence lifetime imaging.
\newblock {\em ACM Transactions on Graphics (TOG)}, 38(3):1--16, 2019.

\bibitem{li2020lidar}
You Li and Javier Ibanez-Guzman.
\newblock Lidar for autonomous driving: The principles, challenges, and trends
  for automotive lidar and perception systems.
\newblock {\em IEEE Signal Processing Magazine}, 37(4):50--61, 2020.

\bibitem{Lindell2018Singlephoton3I}
David~B. Lindell, Matthew O'Toole, and Gordon Wetzstein.
\newblock Single-photon 3d imaging with deep sensor fusion.
\newblock {\em ACM Transactions on Graphics (TOG)}, 37:1 -- 12, 2018.

\bibitem{liu2019non}
Xiaochun Liu, Ib{\'o}n Guill{\'e}n, Marco La~Manna, Ji~Hyun Nam, Syed~Azer
  Reza, Toan~Huu Le, Adrian Jarabo, Diego Gutierrez, and Andreas Velten.
\newblock Non-line-of-sight imaging using phasor-field virtual wave optics.
\newblock {\em Nature}, 572(7771):620--623, 2019.

\bibitem{madonini2021single}
Francesca Madonini and Federica Villa.
\newblock Single photon avalanche diode arrays for time-resolved raman
  spectroscopy.
\newblock {\em Sensors}, 21(13):4287, 2021.

\bibitem{morimoto2020megapixel}
Kazuhiro Morimoto, Andrei Ardelean, Ming-Lo Wu, Arin~Can Ulku, Ivan~Michel
  Antolovic, Claudio Bruschini, and Edoardo Charbon.
\newblock Megapixel time-gated spad image sensor for 2d and 3d imaging
  applications.
\newblock {\em Optica}, 7(4):346--354, 2020.

\bibitem{OToole2018ConfocalNI}
Matthew O'Toole, David~B. Lindell, and Gordon Wetzstein.
\newblock Confocal non-line-of-sight imaging based on the light-cone transform.
\newblock {\em Nature}, 555:338--341, 2018.

\bibitem{Pawlikowska2017SinglephotonTI}
Agata~M. Pawlikowska, Abderrahim Halimi, Robert~A. Lamb, and Gerald~S. Buller.
\newblock Single-photon three-dimensional imaging at up to 10 kilometers range.
\newblock {\em Optics express}, 25 10:11919--11931, 2017.

\bibitem{pediredla2019ellipsoidal}
Adithya Pediredla, Ashok Veeraraghavan, and Ioannis Gkioulekas.
\newblock Ellipsoidal path connections for time-gated rendering.
\newblock {\em ACM Transactions on Graphics (TOG)}, 38(4):1--12, 2019.

\bibitem{Pediredla2018SignalPB}
Adithya~Kumar Pediredla, Aswin~C. Sankaranarayanan, Mauro Buttafava, Alberto
  Tosi, and Ashok Veeraraghavan.
\newblock Signal processing based pile-up compensation for gated single-photon
  avalanche diodes.
\newblock {\em arXiv: Instrumentation and Detectors}, 2018.

\bibitem{pifferi2016new}
Antonio Pifferi, Davide Contini, Alberto Dalla~Mora, Andrea Farina, Lorenzo
  Spinelli, and Alessandro Torricelli.
\newblock New frontiers in time-domain diffuse optics, a review.
\newblock {\em Journal of biomedical optics}, 21(9):091310, 2016.

\bibitem{pittaluga2020towards}
Francesco Pittaluga, Zaid Tasneem, Justin Folden, Brevin Tilmon, Ayan
  Chakrabarti, and Sanjeev~J Koppal.
\newblock Towards a mems-based adaptive lidar.
\newblock In {\em 2020 International Conference on 3D Vision (3DV)}, pages
  1216--1226. IEEE, 2020.

\bibitem{ProjectWebsite}
Ryan Po, Adithya Pediredla, and Ioannis Gkioulekas.
\newblock Project website, 2022.
\newblock \url{https://imaging.cs.cmu.edu/adaptive_gating}.

\bibitem{Rapp2019DeadTC}
Joshua Rapp, Yanting Ma, Robin M.~A. Dawson, and Vivek~K Goyal.
\newblock Dead time compensation for high-flux ranging.
\newblock {\em IEEE Transactions on Signal Processing}, 67:3471--3486, 2019.

\bibitem{RappHighFlux}
Joshua Rapp, Yanting Ma, Robin M.~A. Dawson, and Vivek~K Goyal.
\newblock High-flux single-photon lidar.
\newblock {\em Optica}, 8(1):30--39, Jan 2021.

\bibitem{RappLidarAV}
Joshua Rapp, Julian Tachella, Yoann Altmann, Stephen McLaughlin, and Vivek~K
  Goyal.
\newblock Advances in single-photon lidar for autonomous vehicles: Working
  principles, challenges, and recent advances.
\newblock {\em IEEE Signal Processing Magazine}, 37(4):62--71, 2020.

\bibitem{Rochas2003SinglePA}
Alexis Rochas.
\newblock Single photon avalanche diodes in cmos technology.
\newblock 2003.

\bibitem{russo2017tutorial}
Daniel Russo, Benjamin Van~Roy, Abbas Kazerouni, Ian Osband, and Zheng Wen.
\newblock {A tutorial on Thompson sampling}.
\newblock {\em arXiv preprint arXiv:1707.02038}, 2017.

\bibitem{tadano2015depth}
Ryuichi Tadano, Adithya~Kumar Pediredla, and Ashok Veeraraghavan.
\newblock Depth selective camera: A direct, on-chip, programmable technique for
  depth selectivity in photography.
\newblock In {\em Proceedings of the IEEE International Conference on Computer
  Vision}, pages 3595--3603, 2015.

\bibitem{thompson1933likelihood}
William~R Thompson.
\newblock On the likelihood that one unknown probability exceeds another in
  view of the evidence of two samples.
\newblock {\em Biometrika}, 25(3/4):285--294, 1933.

\bibitem{van2005fluorescence}
Erik~B van Munster and Theodorus~WJ Gadella.
\newblock Fluorescence lifetime imaging microscopy (flim).
\newblock {\em Microscopy techniques}, pages 143--175, 2005.

\bibitem{webb2012beginning}
Jarrett Webb and James Ashley.
\newblock {\em Beginning kinect programming with the microsoft kinect SDK}.
\newblock Apress, 2012.

\bibitem{whyte2015application}
Refael Whyte, Lee Streeter, Michael~J Cree, and Adrian~A Dorrington.
\newblock Application of lidar techniques to time-of-flight range imaging.
\newblock {\em Applied optics}, 54(33):9654--9664, 2015.

\bibitem{Xia2020GeneratingAE}
Zhihao Xia, Patrick Sullivan, and Ayan Chakrabarti.
\newblock Generating and exploiting probabilistic monocular depth estimates.
\newblock {\em 2020 IEEE/CVF Conference on Computer Vision and Pattern
  Recognition (CVPR)}, pages 62--71, 2020.

\bibitem{Xin2019ATO}
Shumian Xin, Sotiris Nousias, Kiriakos~N. Kutulakos, Aswin~C. Sankaranarayanan,
  Srinivasa~G. Narasimhan, and Ioannis Gkioulekas.
\newblock A theory of fermat paths for non-line-of-sight shape reconstruction.
\newblock {\em 2019 IEEE/CVF Conference on Computer Vision and Pattern
  Recognition (CVPR)}, pages 6793--6802, 2019.

\bibitem{yamamoto2018efficient}
Taiki Yamamoto, Yasutomo Kawanishi, Ichiro Ide, Hiroshi Murase, Fumito
  Shinmura, and Daisuke Deguchi.
\newblock Efficient pedestrian scanning by active scan lidar.
\newblock In {\em 2018 International Workshop on Advanced Image Technology
  (IWAIT)}, pages 1--4. IEEE, 2018.

\bibitem{zhao2021high}
Yongyi Zhao, Ankit Raghuram, Hyun Kim, Andreas Hielscher, Jacob~T Robinson, and
  Ashok~Narayanan Veeraraghavan.
\newblock High resolution, deep imaging using confocal time-of-flight diffuse
  optical tomography.
\newblock {\em IEEE Transactions on Pattern Analysis and Machine Intelligence},
  2021.

\end{thebibliography}


\begin{thebibliography}{1}\itemsep=-1pt

\bibitem{Gupta2019AsynchronousS3}
Anant Gupta, Atul Ingle, and Mohit Gupta.
\newblock Asynchronous single-photon 3d imaging.
\newblock {\em 2019 IEEE/CVF International Conference on Computer Vision
  (ICCV)}, pages 7908--7917, 2019.

\bibitem{Gupta2019PhotonFloodedS3}
Anant Gupta, Atul Ingle, Andreas Velten, and Mohit Gupta.
\newblock Photon-flooded single-photon 3d cameras.
\newblock {\em 2019 IEEE/CVF Conference on Computer Vision and Pattern
  Recognition (CVPR)}, pages 6763--6772, 2019.

\bibitem{Heide2018SubpicosecondP3}
Felix Heide, Steven Diamond, David~B. Lindell, and Gordon Wetzstein.
\newblock Sub-picosecond photon-efficient 3d imaging using single-photon
  sensors.
\newblock {\em Scientific Reports}, 8, 2018.

\bibitem{Hernandez2017ACM}
Quercus Hernandez, Diego Gutierrez, and Adri{\'a}n Jarabo.
\newblock A computational model of a single-photon avalanche diode sensor for
  transient imaging.
\newblock {\em ArXiv}, abs/1703.02635, 2017.

\bibitem{liu2019neural}
Chao Liu, Jinwei Gu, Kihwan Kim, Srinivasa~G Narasimhan, and Jan Kautz.
\newblock Neural rgb (r) d sensing: Depth and uncertainty from a video camera.
\newblock In {\em Proceedings of the IEEE/CVF Conference on Computer Vision and
  Pattern Recognition}, pages 10986--10995, 2019.

\bibitem{raaj2021exploiting}
Yaadhav Raaj, Siddharth Ancha, Robert Tamburo, David Held, and Srinivasa~G
  Narasimhan.
\newblock Exploiting \& refining depth distributions with triangulation light
  curtains.
\newblock In {\em Proceedings of the IEEE/CVF Conference on Computer Vision and
  Pattern Recognition}, pages 7434--7442, 2021.

\bibitem{wahl2014time}
Michael Wahl and Sandra Orthaus-M{\"u}ller.
\newblock Time tagged time-resolved fluorescence data collection in life
  sciences.
\newblock {\em Technical Note. PicoQuant GmbH, Germany}, 2014.

\bibitem{Xia2020GeneratingAE}
Zhihao Xia, Patrick Sullivan, and Ayan Chakrabarti.
\newblock Generating and exploiting probabilistic monocular depth estimates.
\newblock {\em 2020 IEEE/CVF Conference on Computer Vision and Pattern
  Recognition (CVPR)}, pages 62--71, 2020.

\end{thebibliography}
}

\end{document}


\title{Adaptive Gating for Single-Photon 3D Imaging: Supplementary Material}

\maketitle

\tableofcontents

\section{Potential negative societal impact}

Our proposed depth sensing system and algorithm are similar to existing LiDAR systems, and thus share similar potential negative societal impacts. In particular, when depth outputs from LiDAR are used to guide decision-making in artificial intelligence systems, such as autonomous vehicles, it is important to account for situations where LiDAR does not perform well. Examples of such situations include bad weather conditions (e.g., snow and rain), strong ambient light (which our work mitigates), and scene elements with very low reflectivity. In the latter situation, it is important to draw attention to the fact that LiDAR performance can vary for different skin types: In particular, darker skin colors reflect less light, and thus result in lower signal to background ratio.

Additionally, LiDAR systems employ high-power lasers, which have associated safety hazards (e.g., risk of blindness). These are typically mitigated using short-wavelength infrared laser wavelengths, where safe power limits are a lot higher. We note that, in our experiments, we used a visible laser due to equipment constraints (we did not have available a laser or SPAD operating at the short-wavelength infrared range); but this is not an inherent limitation of our algorithm.

\section{Timestamp versus histogram measurements}

Prior work on SPAD-based LiDAR~\cite{Gupta2019AsynchronousS3,Gupta2019PhotonFloodedS3,Heide2018SubpicosecondP3} typically assumes that the LiDAR system returns not a detection sequence $\Timestamps$, but a \emph{\detected} $\hsdetected\bracket{\bin}\in\N,\, \bin\in\curly{0,\dots,\Bins-1}$: each histogram entry is equal to the number of photons detected during the corresponding temporal bin after $\numcycles$ cycles. Given the gate sequence $\Gates$, it is easy to convert the detection sequence $\Timestamps$ into the \detected\ $\hsdetected\bracket{\bin}$, but the reverse conversion is not possible. Additionally, the probabilistic models for the detection sequence $\Timestamps$ in Section 3.1 of the main paper and \detected\ $\hsdetected\bracket{\bin}$ in Gupta et al.~\cite{Gupta2019AsynchronousS3} are consistent with each other, assuming that detection times are conditionally independent given gating times. Both our proposed adaptive gating scheme in Section 4 of the main paper and the uniform gating scheme of Gupta et al.~\cite{Gupta2019AsynchronousS3} make use of this conditional independence condition. 

We opt to use detection times in our presentation, as this simplifies the discussion of our adaptive gating scheme. We also note that modern TCSPC systems provide access to individual detection times~\cite{wahl2014time}. However, our adaptive gating scheme would be applicable even if only measurements of the \detected\ were available. In that case, Algorithm 1 of the main paper would use a depth posterior conditioned on the \detected\ $\hsdetected\bracket{\bin}$ instead of the detection sequence $\Timestamp$. Such a posterior could be computed (and updated) using the Poisson-multinomial distribution likelihood model for the \detected\ $\hsdetected\bracket{\bin}$ from Gupta et al.~\cite{Gupta2019AsynchronousS3}.

\section{Assumptions in probabilistic model}

The probabilistic model of Section 3.1 of the main paper makes three important assumptions, also implicit in the prior work we derive this model from~\cite{Gupta2019PhotonFloodedS3,Gupta2019AsynchronousS3}. First, it assumes that the laser emits pulses of infinitesimal duration. Second, it assumes that the TCSPC operation is free from electronic timing noise (\emph{jitter}) and other non-idealities such as afterpulsing~\cite{Hernandez2017ACM}. Third, it assumes that the \transient\ $\rate_\truedepth\bracket{\bin}$ is due to only direct illumination, and is free from indirect illumination effects such as interreflections and subsurface scattering.

We note that we use the infinitesimal-pulse and no-jitter assumptions to simplify exposition, and that they are not essential for our analysis: We can easily adapt the algorithms we introduce in Sections 4-5 of the main paper to use a probabilistic model that replaces the \transient\ $\rate_\truedepth\bracket{\bin}$ of Equation 2 of the main paper with a version that is convolved with arbitrary pulse and jitter shapes. This modification does not increase algorithm complexity, and is analogous to the approach of Heide et al.~\cite{Heide2018SubpicosecondP3} for accounting for finite pulses. Accounting for other TCSPC non-idealities and for indirect illumination is more difficult; we leave these as future research directions.

\section{Alternative reward functions}

We note that an alternative to the reward function of Equation (15) of the main paper is
\begin{equation}
    \label{eq:rewardalt}
    \reward^\ast\paren{\hypotdepth, \gate} \equiv -\Exp{\timestamp\sim\tslikelihood{\timestamp}{\gate}{\hypotdepth}}{\loss\paren{\mapdepth\paren{\Timestamps_\cycle,\Gates_\cycle}, \hypotdepth}}.
\end{equation}
That is, we compute a reward using a depth estimate from not just the next detection time $\timestamp_\cycle$, but also from all already recorded detection times $\Timestamps_{\cycle-1}$. We do not use this reward function for two reasons: First, it increases the computational complexity of Algorithm 1 of the main paper, as we can no longer use Proposition 1 of the main paper to analytically select the gate $\gate_\cycle$ that maximizes $\reward^\ast$; instead, this maximization must be done numerically by exhaustively evaluating the reward function for all possible gate values. Second, we found empirically that using this reward function results in less effective exploration during the early stages of Thompson sampling.


\section{Proof of Proposition 1}

We prove Proposition 1 of the main paper, which we restate here for convenience.

\begin{prop} 
    \label{prop:gate}
    The solution to the optimization problem
    \begin{equation}
        \tilde{\gate} \equiv \argmax_{\gate\in\curly{0,\dots,\Bins-1}} \reward\paren{\hypotdepth, g},
    \end{equation}
    for the reward function 
    \begin{equation}
        \label{eq:reward}
        \reward\paren{\hypotdepth, \gate} \equiv -\Exp{\timestamp\sim\tslikelihood{\timestamp}{\gate}{\hypotdepth}}{\loss\paren{\mapdepth\paren{\timestamp,\gate}, \hypotdepth}},
    \end{equation}
    equals $\tilde{\gate} = \hypotdepth$.
\end{prop}

\begin{proof}
    We note first that, for a single measurement $\timestamp$, $\mapdepth\paren{\timestamp,\gate}=\timestamp \bmod \Bins$. Then,
    \begin{align}
        \reward\paren{\hypotdepth, \gate} &\equiv -\Exp{\timestamp\sim\tslikelihood{\timestamp}{\gate}{\hypotdepth}}{\loss\paren{\mapdepth\paren{\timestamp,\gate}, \hypotdepth}} \\ 
        &= \sum_{\timestamp=\gate}^{\gate+\Bins-1} \tslikelihood{\timestamp}{\gate}{\hypotdepth} \paren{1 - \delta_{\hypotdepth,\timestamp \bmod \Bins}} \\
        &= \sum_{\timestamp=\gate}^{\gate+\Bins-1} \tslikelihood{\timestamp}{\gate}{\hypotdepth} - \tslikelihood{\hypotdepth}{\gate}{\hypotdepth} \\
        &= 1 - \tslikelihood{\hypotdepth}{\gate}{\hypotdepth},
    \end{align}
    %
    where in the last step, we used our assumption that the SPAD always detects a photon during a period equal to the laser's pulse-to-pulse period after the gate (Section 3 of the main paper). This is not an essential assumption, and removing it will simply change the constant $1$ to a number smaller than $1$. Additionally, when we write $\tslikelihood{\hypotdepth}{\gate}{\hypotdepth}$, we slightly abuse our notation to simplify exposition: The first argument should be $\hypotdepth$ if $\hypotdepth > \gate$, and $\hypotdepth + \Bins$ otherwise, to account for the fact that $\timestamp\in\curly{\gate, \gate + \Bins -1}$. 

    Thus,
    \begin{align}
        \tilde{\gate} &\equiv \argmax_{\gate\in\curly{0,\dots,\Bins-1}} \reward\paren{\hypotdepth, g} \\
        &= \argmin_{\gate\in\curly{0,\dots,\Bins-1}} \tslikelihood{\hypotdepth}{\gate}{\hypotdepth} \\
        &= \hypotdepth,
    \end{align}
    where the last step follows from inspection of Equation (5) of the main paper.
\end{proof}
\section{Code and data}

We provide anonymized code and data to reproduce select figures in the paper at the following link: \url{https://anonymous.4open.science/r/adaptive_gating-D53A/README.md}.

Link to data for the ``Leaf" scene can be found here: \url{https://www.dropbox.com/s/y7tx0wafwy6a751/leaf.npy?dl=0}

We will release all of our data upon acceptance. We cannot provide the full scan data as part of the supplement, because of the size limitations for supplementary material.
\section{Experimental details}\label{supsec:details}

\begin{figure}[H]
\centering
    \includegraphics[width=\linewidth]{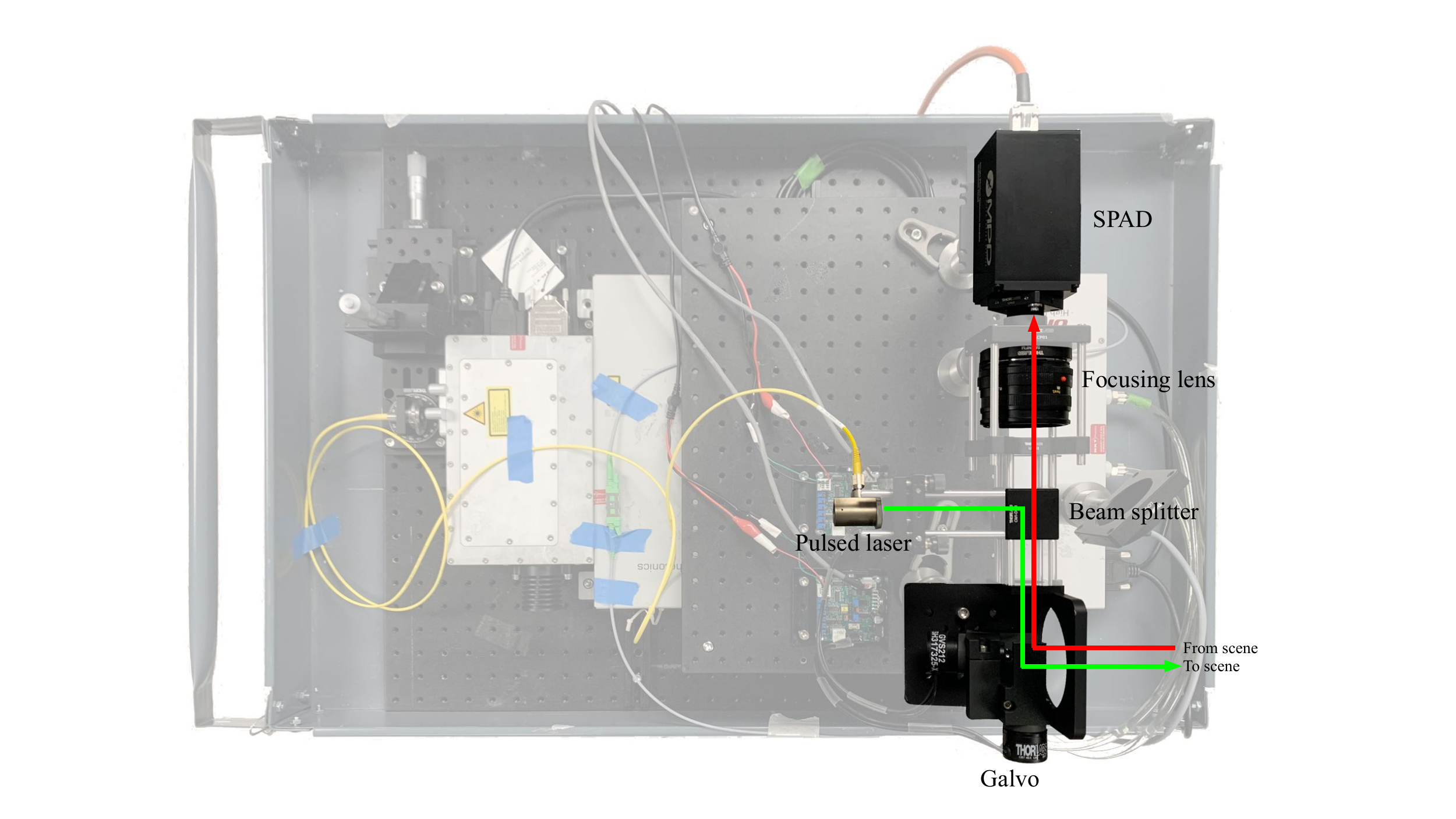}
    \caption{\textbf{Top-down view of experimental setup.} Our SPAD-based LiDAR prototype comprises a fast-gated SPAD and a high-power $\unit[532]{nm}$ picosecond pulsed laser. 
    The setup uses a TCSPC module (not shown) for timestampling photon detections, and a picosecond delayer (not shown) to implement gating. Lastly, the setup uses a pair of galvo mirrors for beam steering.}
    \label{supfig:setup}
\end{figure}

\begin{figure}[H]
\centering
    \includegraphics[width=0.4\linewidth]{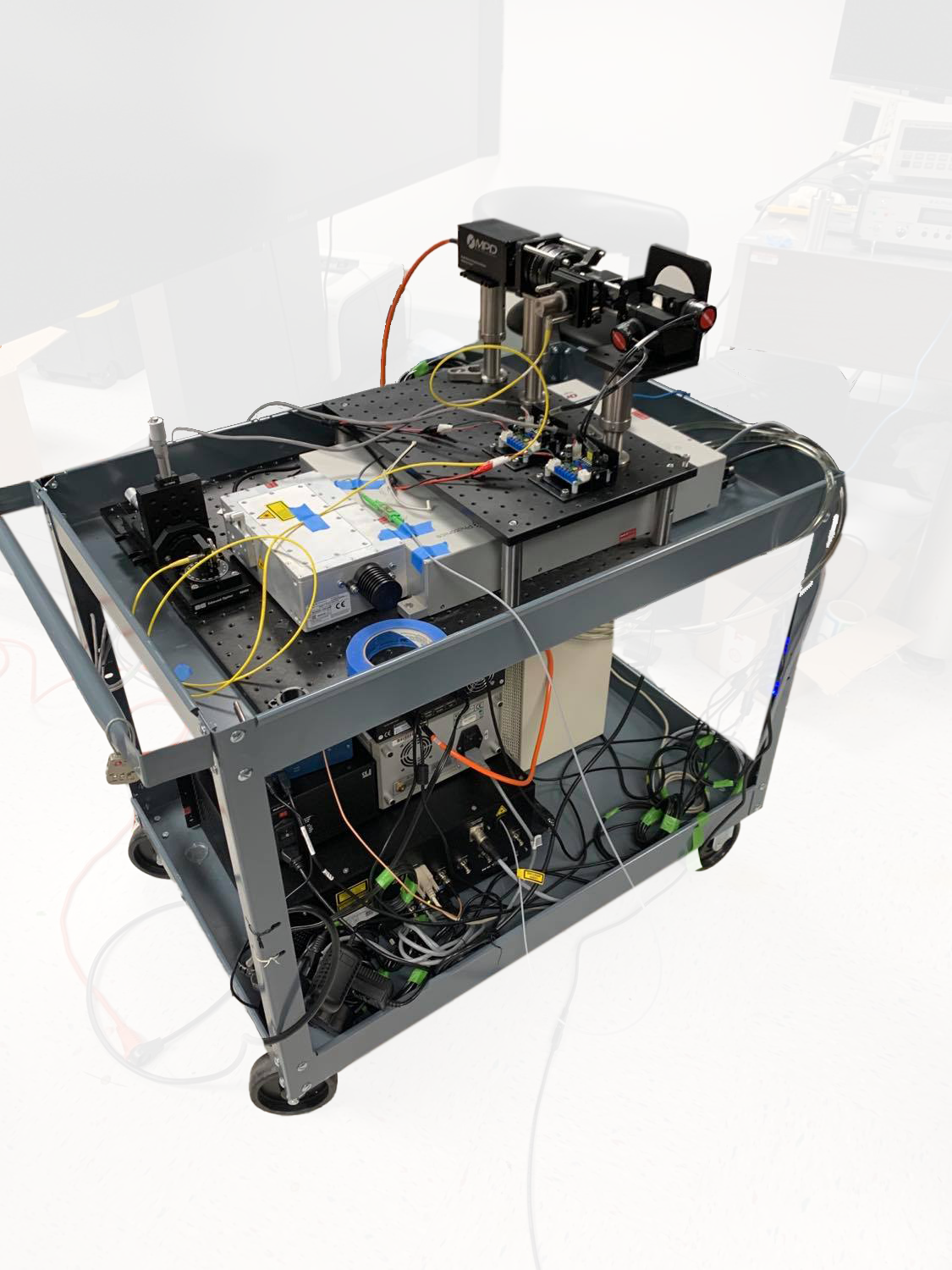}
    \caption{\textbf{Side view of experimental setup.} We place our experimental setup on a cart, to make it easier to capture outdoor scenes.}
    \label{supfig:setup_cart}
\end{figure}

\boldstart{Prototype.} Our SPAD-based LiDAR prototype comprises a fast-gated SPAD (Micro-photon Devices), a $\unit[532]{nm}$ picosecond pulsed laser (NKT Photonics NP100-201-010), a TCSPC module (PicoHarp 300), a programmable picosecond delayer (Micro-photon Devices), and a pair of galvo mirrors (Thorlabs GVS212) for beam steering. We operated the SPAD in triggered mode (adaptive gating) or free-running mode. Supplementary Figures~\ref{supfig:setup}-\ref{supfig:setup_cart} show photographs of our experimental prototype.

\boldstart{Dead time.} In our experiments, we set the SPAD to operate at a dead time of $\unit[81]{ns}$. This is the default setting of the programmable dead time of the SPAD model we used for our experiments. We note that the minimum programmable dead time for our SPAD is $\unit[48]{ns}$. However, we found that operating the SPAD in free-running mode with at dead times smaller than $\unit[81]{ns}$ led to strong artifacts in the captured histograms and detection timestamps. Therefore, we performed all our experiments using the default setting of $\unit[81]{ns}$.

\boldstart{Time discretization.} We follow Gupta et al.~\cite{Gupta2019AsynchronousS3} and discretize our measurements to $\Bins=500$ temporal bins, corresponding to a temporal resolution of $\unit[100]{ps}$. We set the laser pulse frequency to $\unit[20]{MHz}$.

\boldstart{Depth estimation.} We follow Gupta et al.~\cite{Gupta2019AsynchronousS3} (personal communication) and improve depth estimation using a post-processing temporal dithering step. To this end: First, we use either the Coates' or the MAP depth estimate, as we describe in Section 3.2 of the main paper, to form a depth estimate $\hat{\truedepth}$ at a resolution equal to the temporal bin resolution $\Delta$. Then, we fit a Gaussian to the portion of the Coates' estimate of the scene transient (Equation (9) of the main paper) a few temporal bins around the depth estimate $\hat{\truedepth}$. Lastly, we form a new depth estimate at sub-bin resolution, by selecting the max of the fitted Gaussian.



\section{Exposure time comparisons and adaptive exposure}
\begin{figure}[H]
  \centering
   \includegraphics[width=0.8\linewidth]{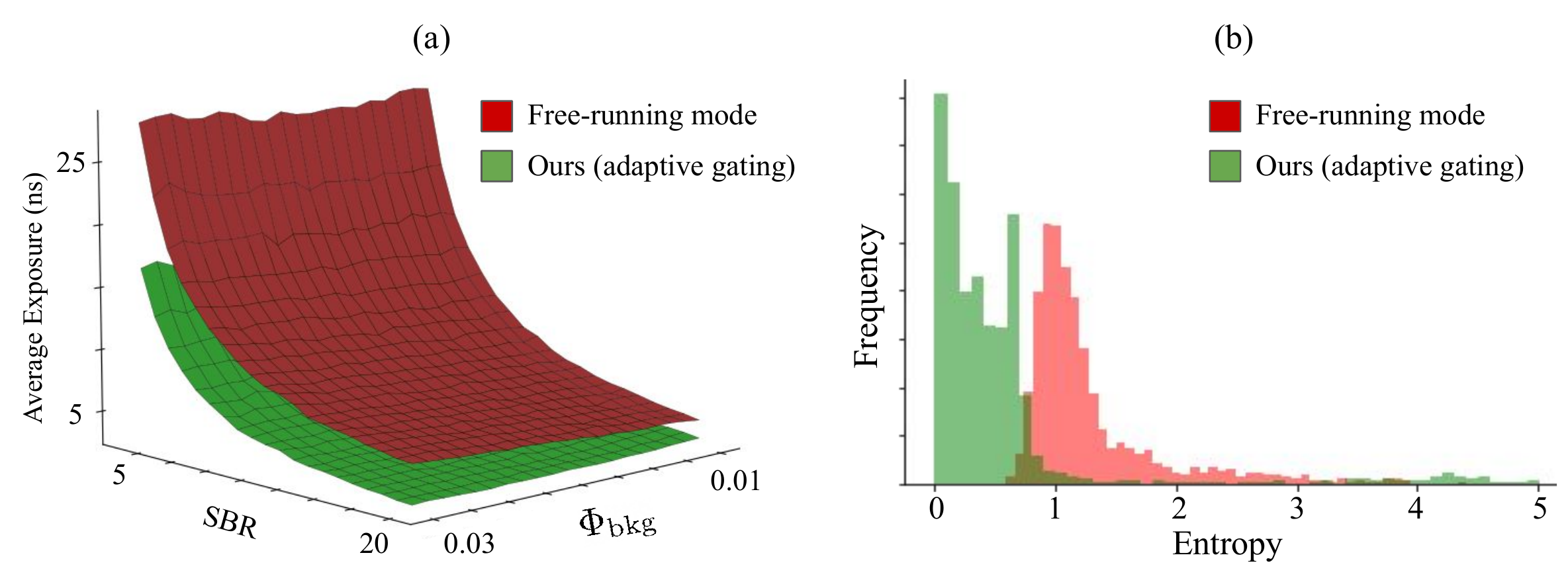}
   \caption{(a) Average exposure for varying levels of ambient flux and SBR using adaptive exposure with an entropy threshold of $0.25$. (b) Distribution of final depth posterior entropy of single-pixel experiment with $\Phi_\text{bkg} \approx 0.01$,  $\Phi_\text{bkg} \approx 0.10$ and a fixed exposure of $100\mu s$.}
   \label{supfig:adaptive_exposure}
\end{figure}
In Supplementary Figure~\ref{supfig:adaptive_exposure}(a), we use simulations to show the exposure time needed by free-running mode and our adaptive gating scheme, to reach a fixed threshold in depth posterior uncertainty (as measured by the termination function), under different SBR and $\bkgflux$ conditions. We note that, in all cases, our adaptive gating scheme requires a shorter exposure time, thanks to the increased number of signal photons it detects. Combining adaptive gating with our adaptive exposure scheme allows us to take advantage of this improved performance automatically, without requiring us to determine a fixed exposure time to be used for each set of SBR and $\bkgflux$ conditions.



In Supplementary Figure~\ref{supfig:adaptive_exposure}(b), we show the depth posterior uncertainty (as measured by termination function) achieved by free-running mode and our adaptive gating scheme under a fixed exposure time, for multiple simulations of the \emph{same} underlying scene transient. We note that, for both algorithms, there is significant variation in the depth posterior uncertainty they achieve, despite the fact that the simulated scene never changes. This is due to the random nature of photon arrivals and, in the case of adaptive gating, the randomness in Thompson sampling. We observe that, in the majority of cases, adaptive gating achieves lower depth posterior uncertainty than free-running mode. We also observe that both algorithms have a ``long tail'' of instances where the achieved depth posterior uncertainty is higher. Our adaptive exposure procedure acts as a simple remedy for such instances: it allows the scanning algorithm to adaptively use longer exposure times in the rare cases when we are in the long tail of this distribution, and use lower exposure times in the majority of cases when we are in the main area of the distribution.

\section{Effect of estimators and exposure time}
\begin{figure}[H]
  \centering
   \includegraphics[width=\linewidth]{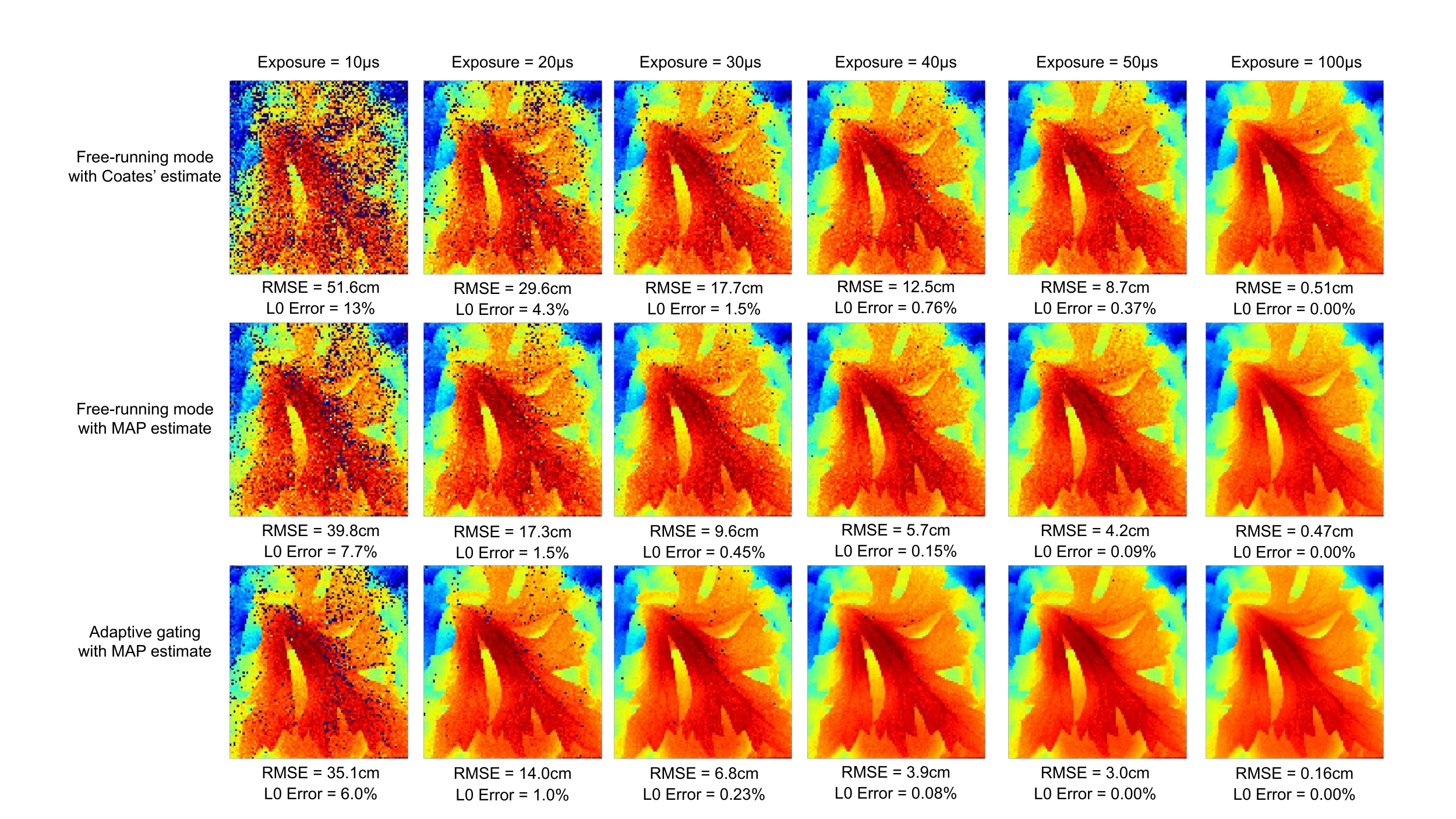}
   \caption{\textbf{Depth reconstruction of ``Leaf" scene under various exposure lengths.} 
   We combine free-running mode with two different estimators, for different exposure times. The combination with the MAP estimator outperforms the combination with Coates' estimator for all exposure times. Additionally, adaptive gating outperforms both versions of free-running mode for all exposure times.}
   \label{supfig:leaf_exposure}
\end{figure} 

As we mention in Section 3.2 of the main paper, our proposed MAP depth estimator can potentially improve depth accuracy compared to the Coates' depth estimator of Gupta et al.~\cite{Gupta2019AsynchronousS3}, independently of how measurement acquisition is performed. To demonstrate this, in the top two rows of Supplementary Figure~\ref{supfig:leaf_exposure}, we use the ``Leaf'' scene to compare the performance of free-running mode when combined with either the Coates' estimator, or our proposed MAP estimator. We note that the MAP estimator consistently outperforms the Coates' estimator. Therefore, in the main paper, for comparisons we always combine free-running mode with the MAP estimator.

We also compare both variants of free-running mode with our adaptive gating scheme (which uses the MAP depth estimator), for different exposure times. We observe that our adaptive gating scheme outperforms both variants of free-running mode under all exposure times.




\section{Effect of dead time}
\begin{figure}[H]
  \centering
   \includegraphics[width=0.45\linewidth]{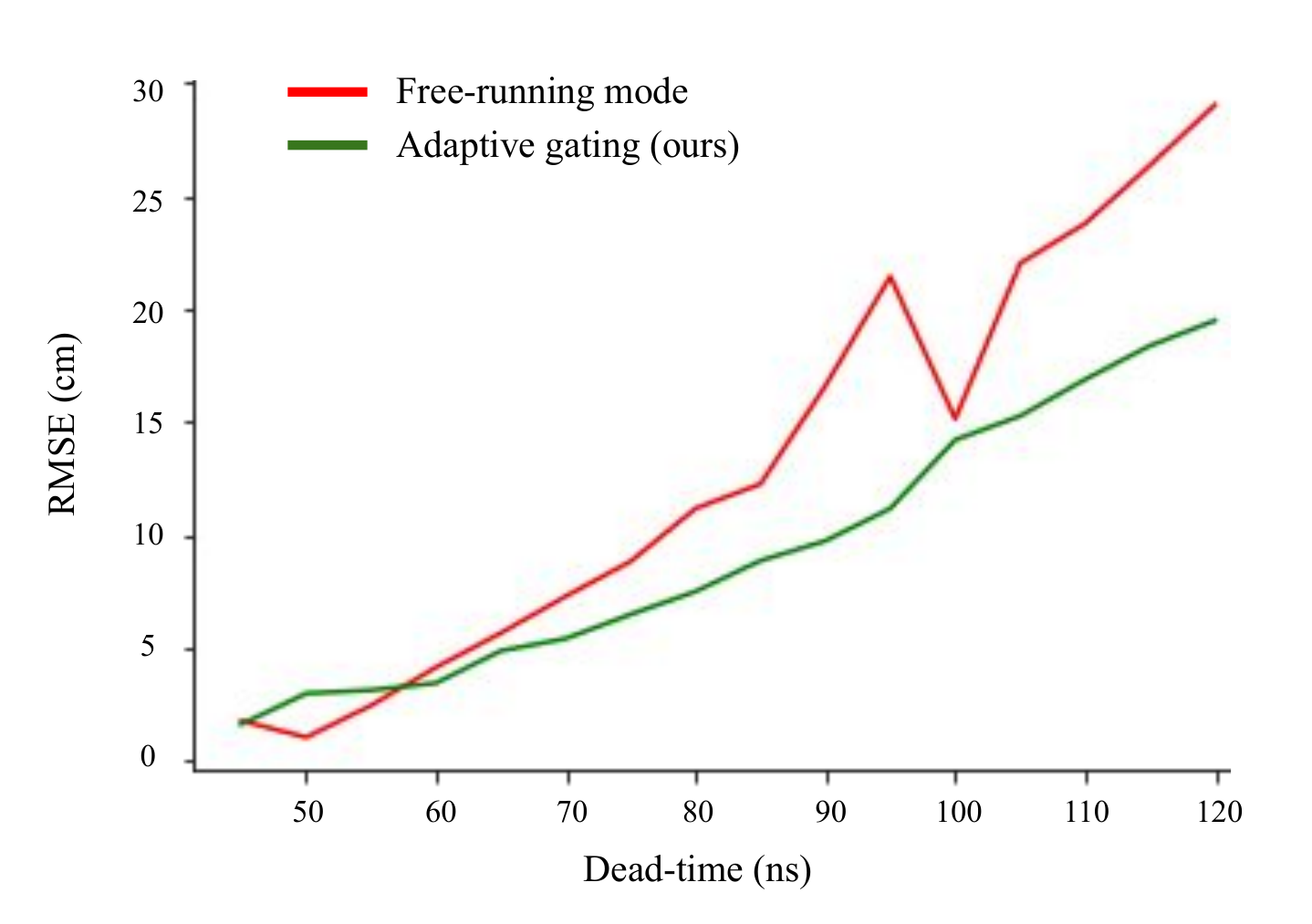}
   \caption{\textbf{Performance of SPAD acquisition modes under various dead time settings.} The performance of adaptive gating degrades as dead time decreases, but remains better than the performance of free-running mode for most dead time settings (including all settings higher than the practical lower bound of $\unit[81]{ns}$). }
   \label{supfig:deadtime}
\end{figure}

Adaptive gating has the undesirable effect of increasing the SPAD inactive time, because it keeps SPAD inactive during the period between the end of dead time in a SPAD cycle and the gate time at the subsequent cycle. By contrast, free-running mode minimizes SPAD inactive time, by activating the SPAD immediately after dead time ends. As a result, free-running mode results in a larger number of photon detections compared to adaptive gating, with this difference increasing as dead time decreases.

In Supplementary Figure~\ref{supfig:deadtime}, we use simulations to evaluate the effect of different dead time settings on depth accuracy, for free-running mode versus our adaptive gating scheme. We note that, even though free-running mode minimizes SPAD inactive time and results in more photon detections, it performs worse than our adaptive gating scheme for most practical dead time settings. As we mention in Supplementary Section~\ref{supsec:details}, we could not reliably operate our SPAD at dead times lower than $\unit[81]{ns}$.

We note an interesting artifact when the dead time equals $\unit[100]{ns}$. At this point, the dead time is equal to the pulse-to-pulse repetition time $\Bins$ that we used for simulations. As a result, free-running mode effectively sets the ``gate'' for the next SPAD cycle at exactly the detection time of the last detected photon. This is exactly analogous to the operation of adaptive gating after the exploration stage is over (i.e., after the first few SPAD cycles), hence explaining the similar performance of the two acquisition schemes at this dead time setting.

It is important to point out here that, if it were possible to reduce dead time to zero, then the SPAD would be able to detect all incident photons, and thus pile-up distortion would not exist. Under such an ideal situation, free-running mode would be the optimal mode of operation for SPADs.
\section{Per-pixel depth uncertainties}
\begin{figure}[H]
  \centering
   \includegraphics[width=\linewidth]{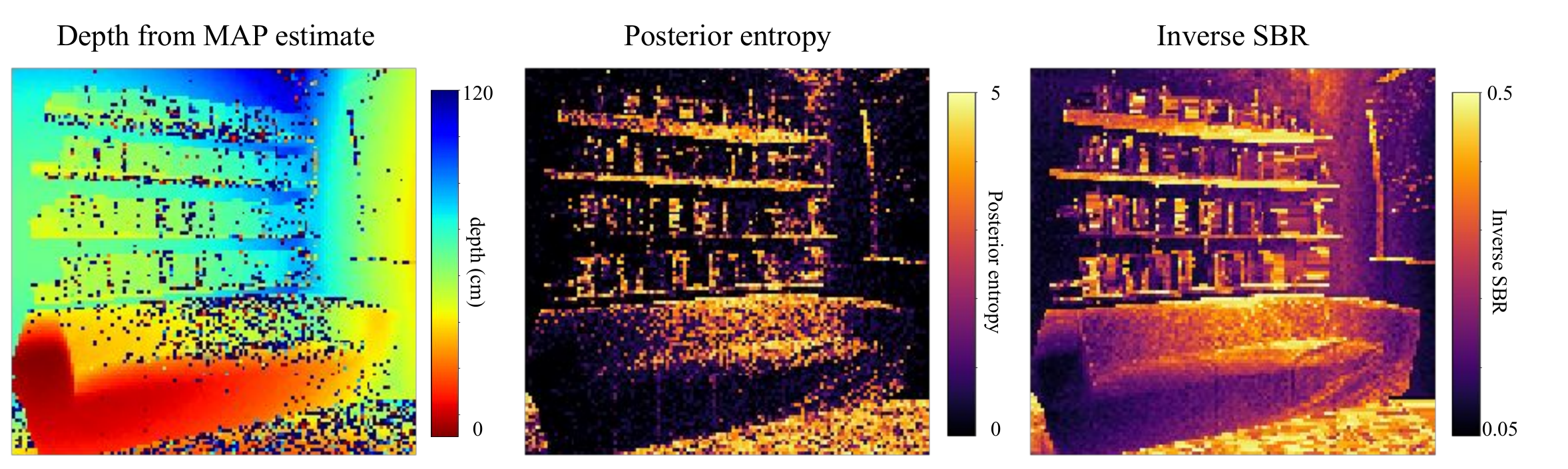}
   \caption{\textbf{Depth reconstruction of ``Office'' scene with corresponding depth posterior entropy at each scene point.} Posterior entropy is high in regions where depth errors occur. As expected, regions with high posterior entropy and high error correspond to regions with high SBR.}
   \label{supfig:office_entropy}
\end{figure}

\begin{figure}[H]
  \centering
   \includegraphics[width=\linewidth]{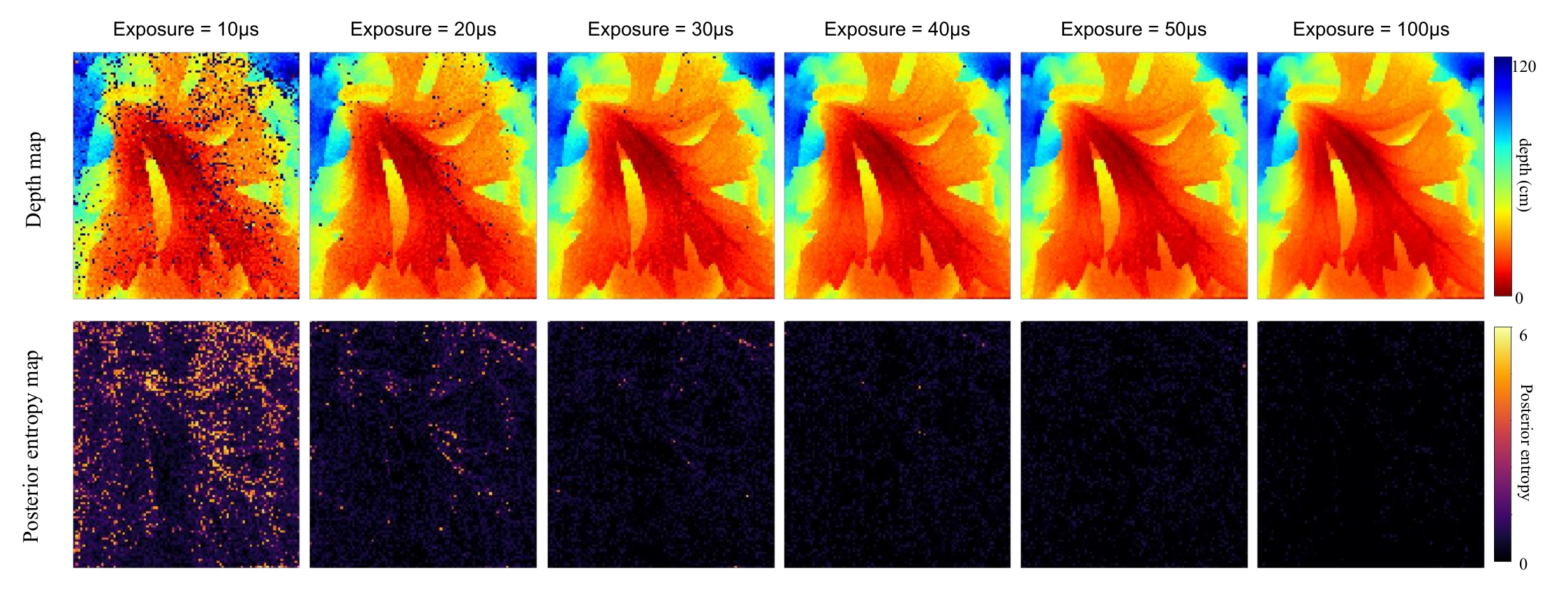}
   \caption{\textbf{Depth reconstruction of ``Leaf'' scene at various exposures with corresponding depth posterior entropy at each scene point.} As exposure increases, the depth posterior entropy at each scene point decreases on average. Regions with high posterior entropy correspond to regions in the depth map with high depth errors.}
   \label{supfig:leaf_entropy}
\end{figure}

Our adaptive gating scheme outputs not just a point estimate for depth, but also the entire depth posterior. This can be useful for downstream tasks that require both point estimates \emph{and} uncertainties~\cite{Xia2020GeneratingAE,raaj2021exploiting,liu2019neural}. As a visualization, in Supplementary Figures~\ref{supfig:office_entropy} and~\ref{supfig:leaf_entropy}, we show per-pixel depth uncertainties (computed as the entropy of the depth posterior) for the ``Ofice'' and ``Leaf'' scenes. In Supplementary Figure~\ref{supfig:office_entropy}, we note that posterior entropy (and thus depth uncertainty) is highly correlated with SBR; in particular, as SBR increases, posterior entropy decreases. This is expected, given that higher SBR implies reduced pile-up artifacts. Additionally, in Supplementary Figure~\ref{supfig:leaf_entropy}, we observe that as exposure time increases, depth uncertainty decreases.
\section{Model mismatch}

\begin{figure}[H]
  \centering
   \includegraphics[width=0.8\linewidth]{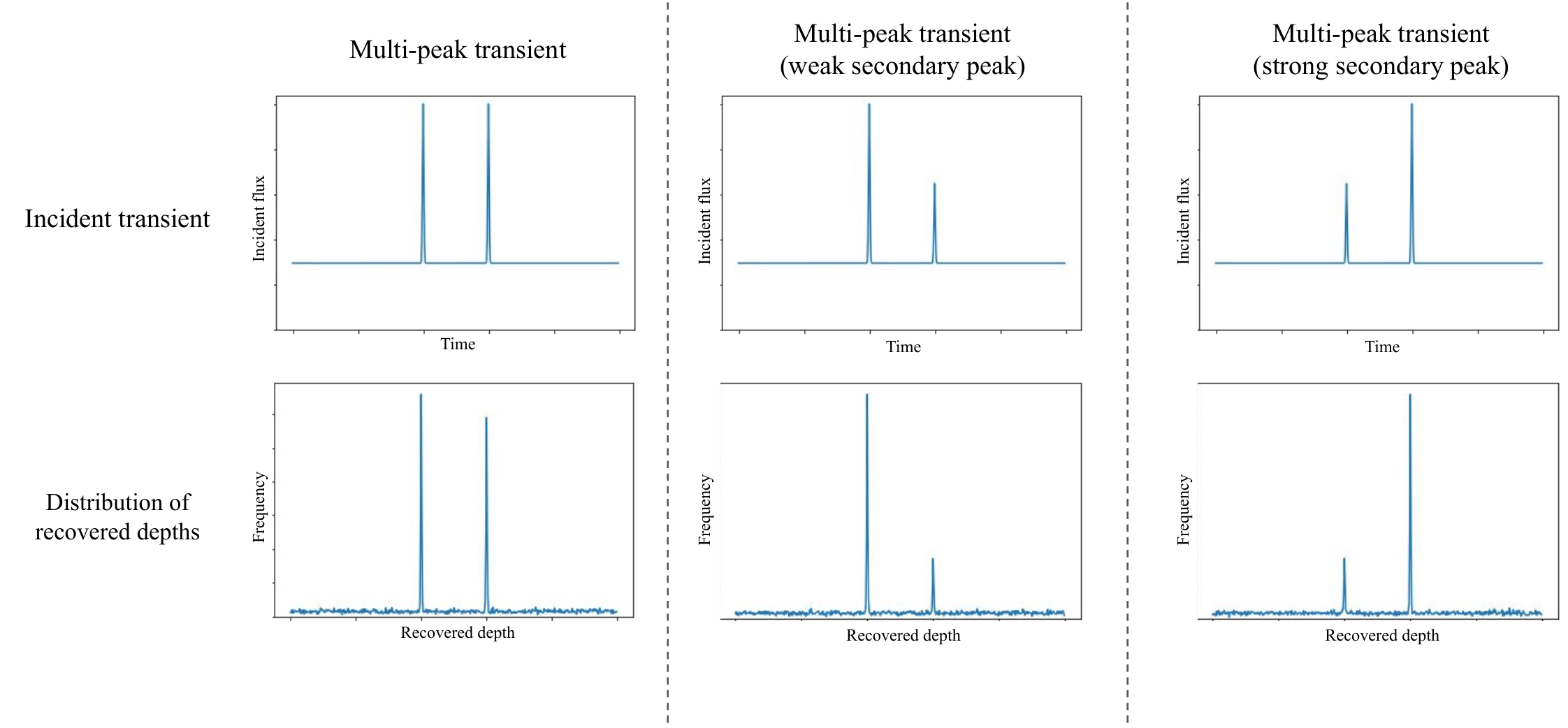}
   \caption{\textbf{Behavior of adaptive gating under model mismatch: multiple peaks.} When multiple peaks are present in the incident transient, a model mismatch occurs to the single-peaked probabilistic model for the MAP estimator. The behavior of adaptive gating under such scenarios depends on the relative strengths of these peaks. Adaptive gating will probabilistically converge to one of these depths, with likelihood positively correlated to the signal strength at that depth.}
   \label{supfig:model_mismatch}
\end{figure}

\begin{figure}[H]
  \centering
   \includegraphics[width=0.8\linewidth]{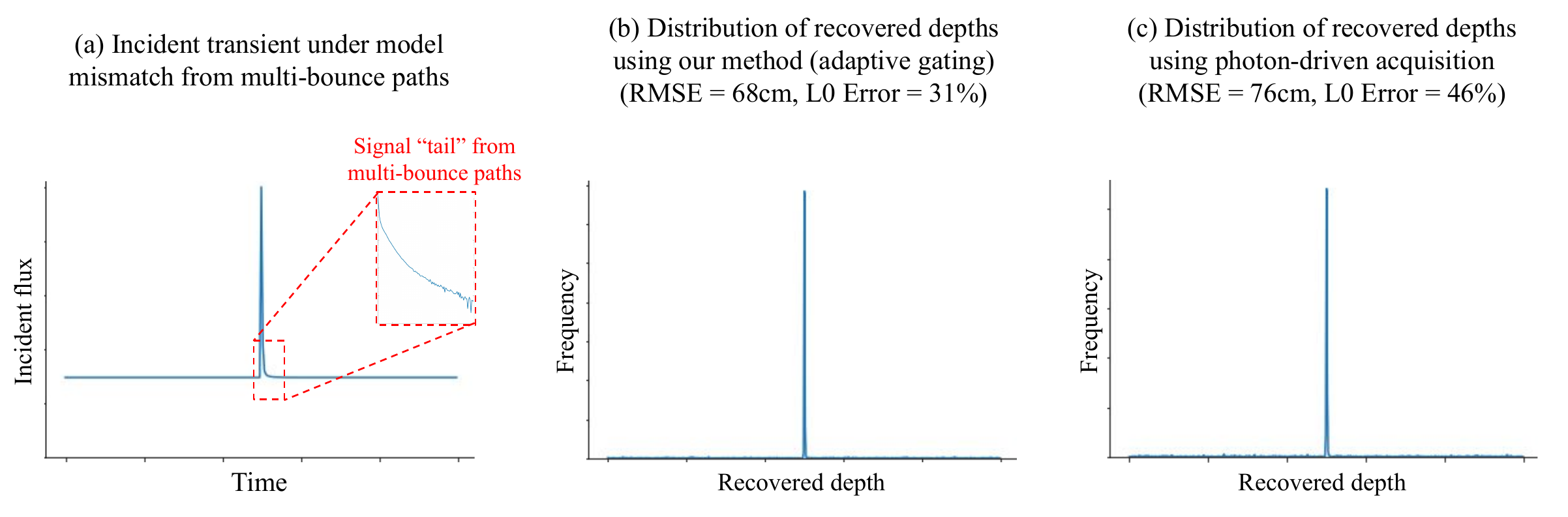}
   \caption{\textbf{Behavior of adaptive gating under model mismatch: multi-bounce pahts.} (a) shows an example transient resulting from imaging a corner, where the presence of mult-bounce paths leads to a ``tail'' appearing after the true signal. (b) and (c) shows the distribution of recovered depths using our method (adaptive gating) and photon-driven acquisition respectively. Note that under model mismatch, our method still performs better than photon-driven acquisition.}
   \label{supfig:multibounce}
\end{figure}

In scenes where multi-bounce paths are common, e.g. a corner, the incident transient may not adhere well to the assumed model. In such cases, multi-bounce paths causes additional photons to return after the initial signal, resulting in a decaying ``tail'' that can be observed in Supplmentary Fig. \ref{supfig:multibounce}(a). Since light returning to the SPAD after experiencing multiple bounces tend to be much weaker and also due to the confocal nature of our setup, multi-bounce artifacts are often very weak, meaning the transients still closely ressemble the assumed model. Supplmentary Fig. \ref{supfig:multibounce}(b) and (c) illustrates how our method and photon-driven method are able to recover the correct depth under model mismatch, with our method still performing better despite deviation from our assumed model.
\section{Additional figures}

\begin{figure}[H]
  \centering
   \includegraphics[width=\linewidth]{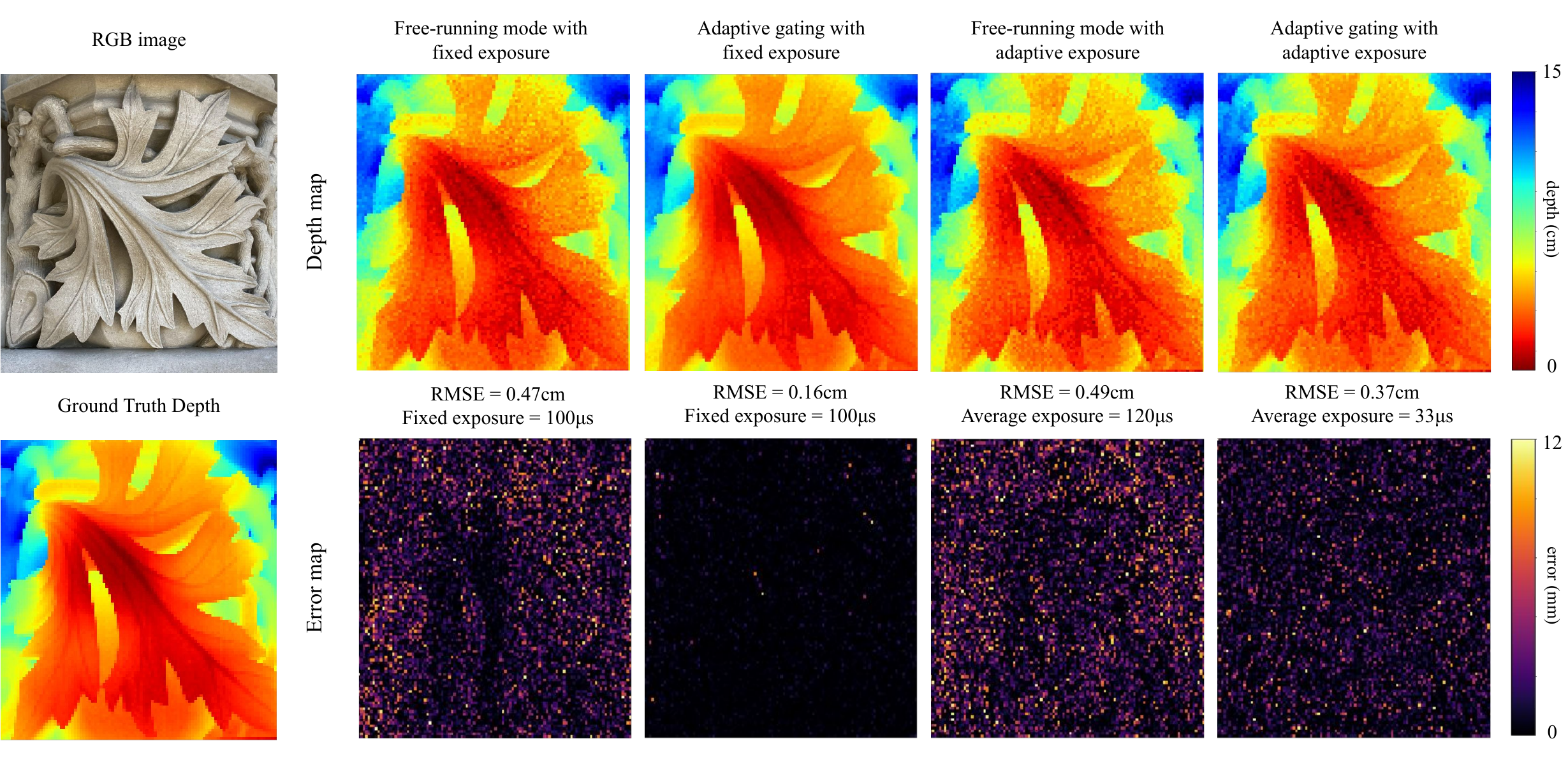}
   \caption{\textbf{Depth reconstruction of ``Leaf'' scene with corresponding error maps.} Under equal fixed integration times, adaptive gating leads to lower errors across the entire scene compared to free-running mode. Using adaptive exposure, adaptive gating is able to achieve lower RMSE despite only using $\frac{1}{3}$ of the total exposure time.}
   \label{supfig:leaf_additional}
\end{figure}

\begin{figure}[H]
  \centering
   \includegraphics[width=0.8\linewidth]{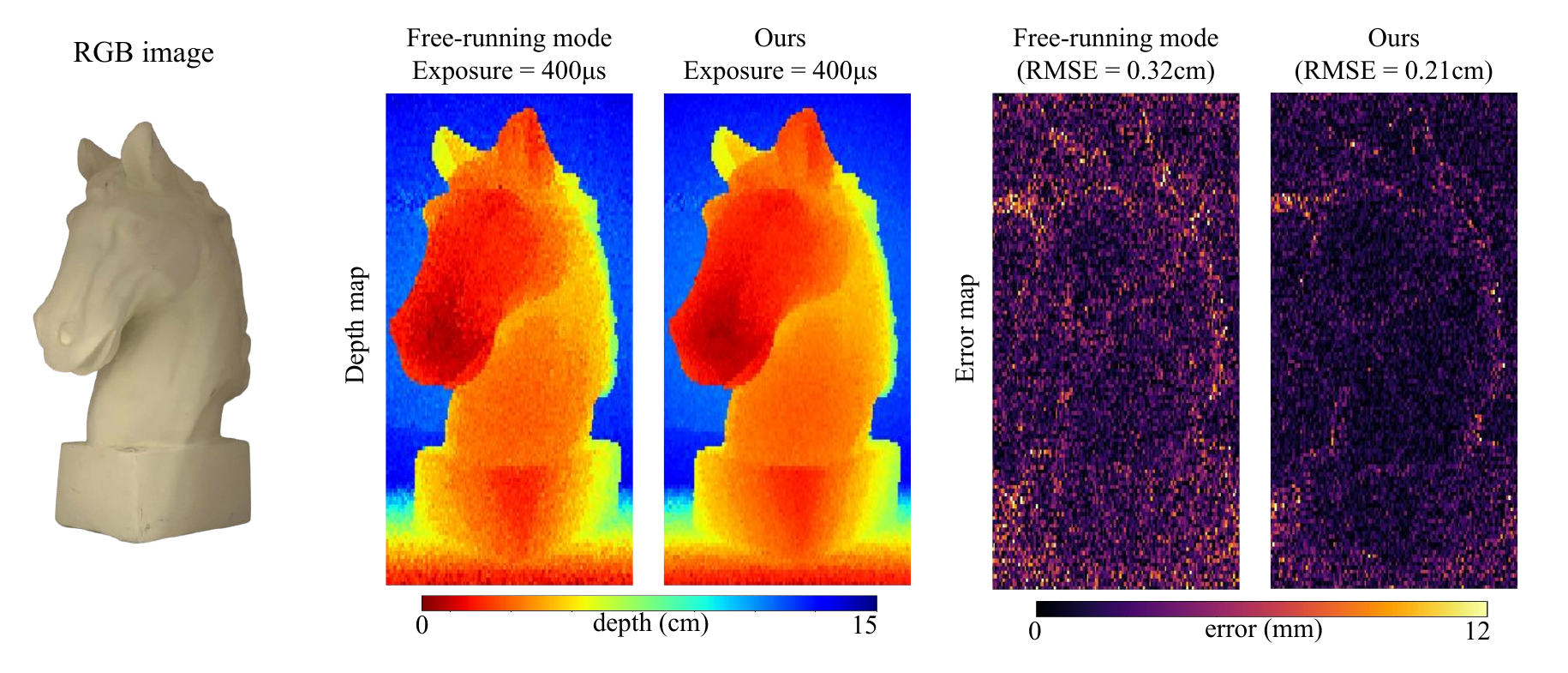}
   \caption{\textbf{Depth reconstruction for ``Horse'' scene with corresponding error maps} Our method gives lower errors across all regions of the image.}
   \label{supfig:horse_additional}
\end{figure}

\begin{figure}[H]
  \centering
   \includegraphics[width=\linewidth]{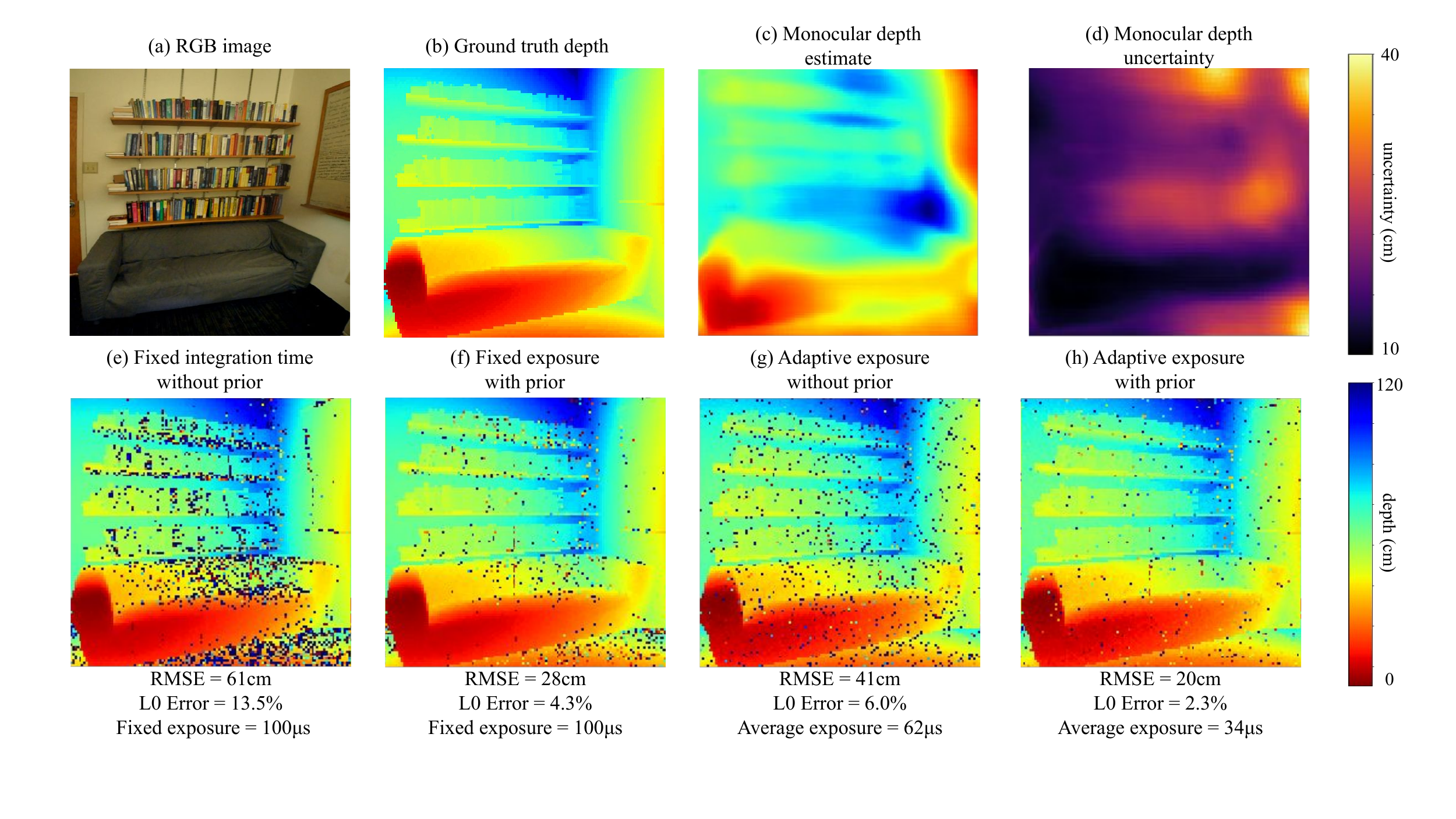}
   \caption{\textbf{Depth reconstruction of ``Office'' using monocular depth priors.} (c) and (d) show the depth estimates and corresponding uncertainties produced by the learning-based monocular depth technique of Xia et al.~\cite{Xia2020GeneratingAE}. We use these results to construct a per-pixel Gaussian depth prior with mean equal to the depth estimates in (c), and variance equal to the uncertainties shown in (d). Using this prior with our adaptive gating scheme leads to significantly better depth reconstructions (e-f). When we additionally use adaptive exposure, using the depth prior also helps decrease total exposure time (g-h).}
   \label{supfig:office_additional}
\end{figure}

Supplementary Figures~\ref{supfig:leaf_additional},~\ref{supfig:horse_additional}, and~\ref{supfig:office_additional} show additional visualizations for the experiments on the ``Leaf'', ``Horse'', and ``Office'' scenes we report in the main paper. We note that we use the indoor scene ``Office'', where pile-up artifacts are limited, to showcase the ability of our adaptive gating scheme to take advantage of priors from the monocular depth technique of Xia et al.~\cite{Xia2020GeneratingAE}. Unfortunately, we were unable to obtain reliable depth and uncertainty estimates from the technique of Xia et al.~\cite{Xia2020GeneratingAE} for outdoor scenes, likely because the neural network it uses is trained on a dataset of indoor scenes.

\begin{figure}[H]
  \centering
   \includegraphics[width=0.8\linewidth]{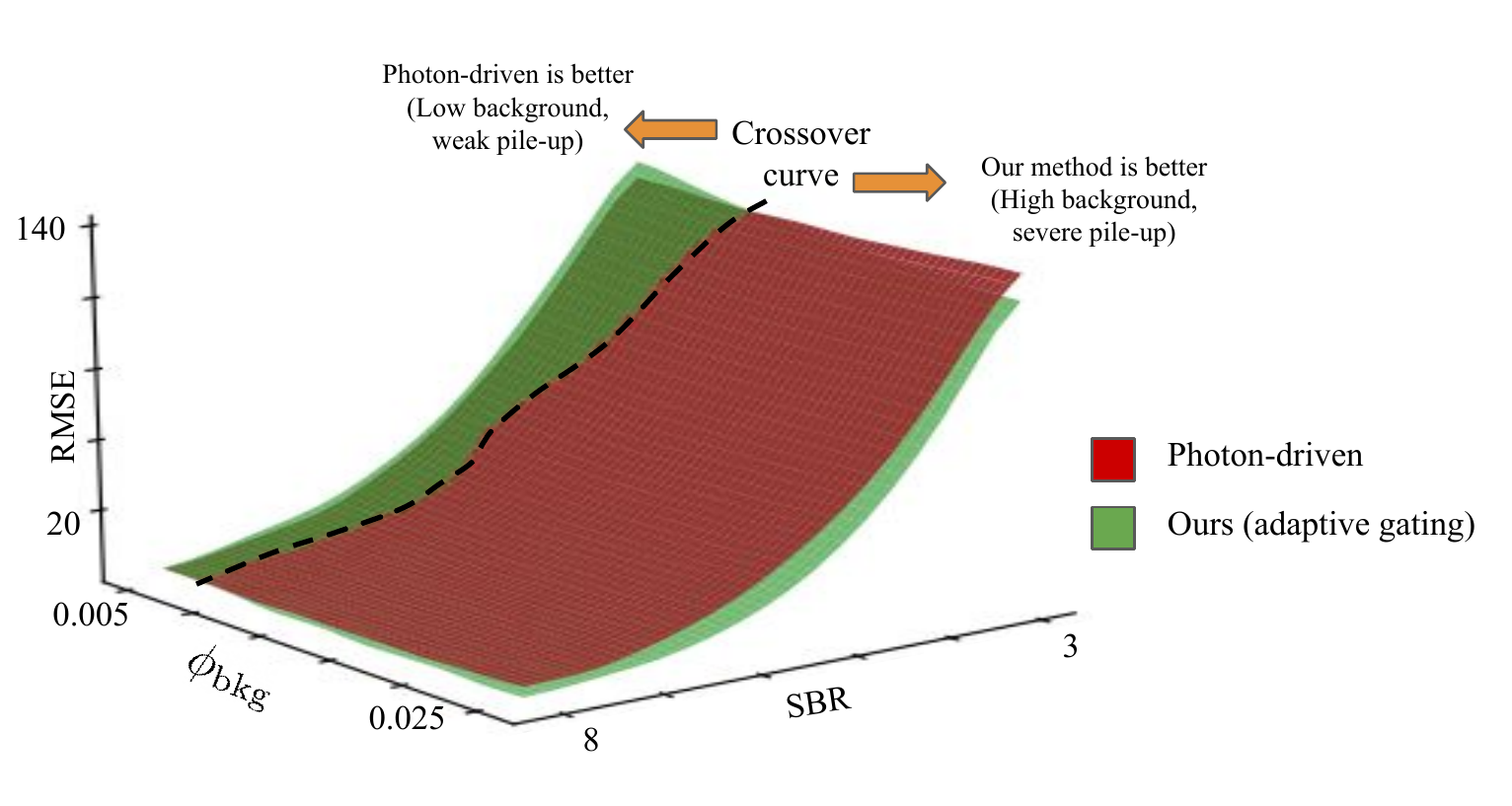}
   \caption{\textbf{Performance comparison between adaptive gating and photon-driven acquisition} Adaptive gating consistently out-performs photon-driven acquisition except in scenarios of low background light. Adaptive gating facilitates better exploration of the depth range when pile-up effects are present, this explains the decrease in RMSE observed when background intensities are high. However, since optimal gate selection leads to additional SPAD dead-time, in low background light scenarios where pile-up effects are no longer observed, photon-driven acquisition performs better than adaptive gating.}
   \label{supfig:rmse_ad_vs_pd}
\end{figure}

{\small
\bibliographystyle{ieee_fullname}
\bibliography{adaptive_supplement}
}